\documentclass{article}

\usepackage[final]{neurips_2022}
\usepackage[utf8]{inputenc}
\usepackage[T1]{fontenc}
\usepackage{hyperref}
\usepackage{url}
\usepackage{booktabs}
\usepackage{amsfonts}
\usepackage{nicefrac}
\usepackage{microtype}
\usepackage{xcolor}
\usepackage{mathtools}
\usepackage{amsmath}
\usepackage{amsthm}
\usepackage{cleveref}
\usepackage{tcolorbox}
\usepackage{placeins}
\usepackage{amssymb}
\usepackage{bm}
\usepackage{float}

\newcommand{\Xm}{\mathbb{X}_m}

\newcommand{\Yn}{\mathbb{Y}_n}
\newcommand{\Zn}{\mathbb{Z}_N}
\newcommand{\D}{\mathcal{D}}

\newcommand{\R}{\mathbb R}
\newcommand{\N}{\mathbb N}
\newcommand{\E}{\mathbb E}
\newcommand{\PP}{\mathbb P}
\newcommand{\abss}[1]{\big|#1\big|}
\newcommand{\p}[1]{\left( #1 \right)}
\newcommand{\pp}[1]{\!\left( #1 \right)}

\newcommand{\pbig}[1]{\big( #1 \big)}

\newcommand{\pBig}[1]{\Big( #1 \Big)}

\newcommand{\Kt}{\tilde{\textbf{K}}}
\newcommand{\Lt}{\tilde{\textbf{L}}}
\newcommand{\one}{\textbf{1}}
\newcommand{\LL}{\lambda_1\cdots\lambda_d}
\newcommand{\LLs}{\lambda_1^*\cdots\lambda_d^*}
\newcommand{\Sb}{\mathcal{S}_d^s(R)}

\DeclarePairedDelimiter\abs{\lvert}{\rvert}
\DeclarePairedDelimiter\norm{\lVert}{\rVert}
\makeatletter
\let\oldabs\abs
\def\abs{\@ifstar{\oldabs}{\oldabs*}}
\let\oldnorm\norm
\def\norm{\@ifstar{\oldnorm}{\oldnorm*}}
\makeatother

\newtheorem{theorem}{Theorem}

\newtheorem{proposition}{Proposition}
\newtheorem{lemma}{Lemma}

\title{Efficient Aggregated Kernel Tests \\ using Incomplete $U$-statistics}

\author{%
Antonin Schrab \\
Centre for Artificial Intelligence\\
Gatsby Computational Neuroscience Unit\\
University College London \& Inria London \\
\texttt{a.schrab@ucl.ac.uk}
\And
Ilmun Kim \\
Department of Statistics \& Data Science\\
Department of Applied Statistics\\
Yonsei University \\ 
\texttt{ilmun@yonsei.ac.kr}
\And
Benjamin Guedj \\
Centre for Artificial Intelligence\\
University College London \& Inria London \\
\texttt{b.guedj@ucl.ac.uk} \\
\And
Arthur Gretton \\
Gatsby Computational Neuroscience Unit\\
University College London \\
\texttt{arthur.gretton@gmail.com}
}

\begin{document}

\maketitle

\begin{abstract}
We propose a series of computationally efficient nonparametric tests for the two-sample, independence, and goodness-of-fit problems, using the  Maximum Mean Discrepancy (MMD), Hilbert Schmidt Independence Criterion (HSIC), and Kernel Stein Discrepancy (KSD), respectively. 
Our test statistics are incomplete $U$-statistics, with a computational cost that interpolates between linear time in the number of samples, and quadratic time, as associated with classical $U$-statistic tests.
The three proposed tests aggregate over several kernel bandwidths to detect departures from the null on various scales: we call the resulting tests MMDAggInc, HSICAggInc and KSDAggInc.
This procedure provides a solution to the fundamental kernel selection problem as we can aggregate a large number of kernels with several bandwidths without incurring a significant loss of test power.
For the test thresholds, we derive a quantile bound for wild bootstrapped incomplete $U$-statistics, which is of independent interest.
We derive non-asymptotic uniform separation rates for MMDAggInc and HSICAggInc, and quantify exactly the trade-off between computational efficiency and the attainable rates: this result is novel for tests based on incomplete $U$-statistics, to our knowledge.
We further show that in the quadratic-time case, the wild bootstrap incurs no penalty to test power over the more widespread permutation-based approach, since both attain the same minimax optimal rates (which in turn match the rates that use oracle quantiles).
We support our claims with numerical experiments on the trade-off between computational efficiency and test power. 
In all three testing frameworks, the linear-time versions of our proposed tests perform at least as well as the current linear-time state-of-the-art tests.
\end{abstract}

\section{Introduction}
\label{sec:introduction}

Nonparametric hypothesis testing is a fundamental field of statistics, and is widely used by the machine learning community and practitioners in numerous other fields, due to the increasing availability of huge amounts of data. When dealing with large-scale datasets, computational cost can quickly emerge as a major issue which might prevent from using expensive tests in practice; constructing efficient tests is therefore crucial for their real-world applications.
In this paper, we construct kernel-based aggregated tests using incomplete $U$-statistics \citep{blom1976some} for the \textbf{two-sample}, \textbf{independence} and \textbf{goodness-of-fit} problems (which we detail in \Cref{sec:background}).
The quadratic-time aggregation procedure has been shown to result in powerful tests (\citealp{fromont2012kernels}; \citealp{fromont2013two}; \citealp{albert2019adaptive}; \citealp{schrab2021mmd,schrab2022ksd}), we propose efficient variants of these well-studied tests, with computational cost interpolating from the classical quadratic-time regime to the linear-time one.

\textbf{Related work: aggregated tests.}
Kernel selection (or kernel bandwidth selection) is a fundamental problem in nonparametric hypothesis testing as this choice has a major influence on test power.
Motivated by this problem, non-asymptotic aggregated tests, which combine tests with different kernel bandwidths, have been proposed for the two-sample \citep{fromont2012kernels,fromont2013two,kim2020minimax,schrab2021mmd}, independence \citep{albert2019adaptive,kim2020minimax}, and goodness-of-fit \citep{schrab2022ksd} testing frameworks.
\citet{li2019optimality} and
\citet{balasubramanian2021optimality} construct similar aggregated tests for these three problems, with the difference that they work in the asymptotic regime.
All the mentioned works study aggregated tests in terms of uniform separation rates \citep{baraud2002non}.
Those rates depend on the sample size and satisfy the following property: if the $L^2$-norm difference between the densities is greater than the uniform separation rate, then the test is guaranteed to have high power.
All aggregated kernel-based tests in the existing literature have been studied using $U$-statistic estimators \citep{hoeffding1992class} with tests running in quadratic time.

\textbf{Related work: efficient kernel tests.}
Several linear-time kernel tests have been proposed for those three testing frameworks.
Those include
tests using classical linear-time estimators with median bandwidth \citep{gretton2012kernel,liu2016kernelized}
or selecting an optimal bandwidth on held-out data to maximize power \citep{gretton2012optimal},
tests using eigenspectrum approximation \citep{gretton2009fast},
tests using post-selection inference for adaptive kernel selection with incomplete $U$-statistics \citep{yamada2018post,yamada2019post,lim2019kernel,lim2020more,kubler2020learning,freidling2021post},
tests which use a Nystr\"om approximation of the asymptotic null distribution \citep{zhang2018large,cherfaoui2022discrete},
random Fourier features tests \citep{zhang2018large,zhao2015fastmmd,chwialkowski2015fast},
tests based on random feature Stein discrepancies \citep{huggins2018random},
the  adaptive tests which use features selected on held-out data to maximize power \citep{jitkrittum2016interpretable,jitkrittum2017adaptive,jitkrittum2017linear}, as well as tests using neural networks to learn a discrepancy \citep{grathwohl2020learning}.
We also point out the very relevant works of \citet{kubler2022witness} on a quadratic-time test, and of \citet{ho2006two}, \citet{zaremba2013b} and \citet{zhang2018large} on the use of block $U$-statistics with complexity $\mathcal{O}(N^{1.5})$ for block size $\sqrt{N}$ where $N$ is the sample size.

\textbf{Contributions and outline.}
In \Cref{sec:background}, we present the three testing problems with their associated well-known quadratic-time kernel-based estimators (MMD, HSIC, KSD) which are $U$-statistics.
We introduce three associated incomplete $U$-statistics estimators, which can be computed efficiently, in \Cref{sec:incomplete}.
We then provide quantile and variance bounds for generic incomplete $U$-statistics using a wild bootstrap, in \Cref{sec:bounds}.
We study the level and power guarantees at every finite sample sizes for our efficient tests using incomplete $U$-statistics for a fixed kernel bandwidth, in \Cref{sec:rates}. 
In particular, we obtain non-asymptotic uniform separation rates for the two-sample and independence tests over a Sobolev ball, and show that these rates are minimax optimal up to the cost incurred for efficiency of the test.
In \Cref{sec:aggregation}, we propose our efficient aggregated tests which combine tests with multiple kernel bandwidths. 
We prove that the proposed tests are adaptive over Sobolev balls and achieve the same uniform separation rate (up to an iterated logarithmic term) as the tests with optimal bandwidths.
As a result of our analysis, we have shown minimax optimality over Sobolev balls of the quadratic-time tests using quantiles estimated with a wild bootstrap.
Whether this optimality result also holds for tests using the more general permutation-based procedure to approximate HSIC quantiles, was an open problem formulated by \citet{kim2020minimax}, we prove that it indeed holds in \Cref{sec:minimaxproof}.
As observed in \Cref{sec:experiments}, the linear-time versions of MMDAggInc, HSICAggInc and KSDAggInc retain high power, and either outperform or match the power of other state-of-the-art linear-time kernel tests.
Our implementation of the tests and code for reproducibility of the experiments are available online under the MIT license:
\url{https://github.com/antoninschrab/agginc-paper}.

\section{Background}
\label{sec:background}

In this section, we briefly describe our main problems of interest, comprising the two-sample, independence and goodness-of-fit problems. We approach these problems from a nonparametric point of view using the kernel-based statistics: MMD, HSIC, and KSD. We briefly introduce original forms of these statistics, which can be computed in quadratic time, and also discuss ways of calibrating tests proposed in the literature. The three quadratic-time expressions are presented in \Cref{sec:detailedbackground}.

\textbf{Two-sample testing.} 
In this problem, we are given independent samples $\Xm \coloneqq (X_i)_{1\leq i\leq m}$ and $\Yn= (Y_j)_{1\leq j\leq n}$, consisting of i.i.d.\ random variables with respective probability density functions\footnote{All probability density functions in this paper are with respect to the Lebesgue measure.}  $p$ and $q$ on $\R^d$. 
We assume we work with balanced sample sizes, that is\footnote{We use the notation $a \lesssim b$ when there exists a constant $C>0$ such that $a\leq C b$. We similarly use the notation $\gtrsim$. We write $a\asymp b$ if $a \lesssim b$ and $a \gtrsim b$.   We also use the convention that all constants are generically denoted by $C$, even though they might be different.} $\max(m,n)\lesssim \min(m,n)$.
We are interested in testing the null hypothesis $\mathcal{H}_0: p=q$ against the alternative $\mathcal{H}_1:p\neq q$; that is, we want to know if the samples come from the same distribution. 
\citet{gretton2012kernel} propose a nonparametric kernel test based on the {\em Maximum Mean Discrepancy} (MMD), a measure between probability distributions which uses a characteristic kernel $k$ \citep{fukumizu2008kernel,sriperumbudur2011universality}.
It can be estimated using a quadratic-time estimator \citep[Lemma~6]{gretton2012kernel} which, as noted by \citet{kim2020minimax}, can be expressed as a two-sample $U$-statistic (both of second order) \citep{hoeffding1992class},
\begin{equation}
    \label{Ummd}
    \widehat{\mathrm{MMD}}^2_k(\Xm,\Yn) = 
    \frac{1}{\abss{\textbf{i}_2^m}\abss{\textbf{i}_2^n}} 
    \sum_{(i,i')\in \textbf{i}_2^m}
    \sum_{(j,j')\in \textbf{i}_2^n}
    h_k^{\mathrm{MMD}}(X_i, X_{i'}; Y_j, Y_{j'}),
\end{equation}
where $\textbf{i}_a^b$ with $a\leq b$ denotes the set of all $a$-tuples drawn without replacement from $\{1,\dots,b\}$ so that $\abss{\textbf{i}_a^b} = b \cdots (b-a+1)$, and where, for $x_1,x_2,y_1,y_2\in\R^d$, we let
\begin{equation}
    \label{h_mmd}
    h_k^{\mathrm{MMD}}(x_1, x_2; y_1, y_2) \coloneqq k(x_1,x_2)  - k(x_1,y_2) - k(x_2,y_1) + k(y_1,y_2).
\end{equation}

\textbf{Independence testing.} 
In this problem, we have access to i.i.d.\ pairs of samples $\Zn \coloneqq \big(Z_i\big)_{1\leq i\leq N} = \big((X_i,Y_i)\big)_{1\leq i\leq N}$ with joint probability density $p_{xy}$ on $\R^{d_x}\times\R^{d_y}$ and marginals $p_x$ on $\R^{d_x}$ and $p_y$ on $\R^{d_y}$. 
We are interested in testing $\mathcal{H}_0: p_{xy}=p_x\otimes p_y$ against $\mathcal{H}_1:p_{xy}\neq p_x\otimes p_y$; that is, we want to know if two components of the pairs of samples are independent or dependent.
\citet{gretton2005kernel,gretton2008kernel} propose a nonparametric kernel test based on the {\em Hilbert Schmidt Independence Criterion} (HSIC).
It can be estimated using the quadratic-time estimator proposed by \citet[Equation 5]{song2012feature} which is a fourth-order one-sample $U$-statistic
\begin{align}
    \label{Uhsic}
    \widehat{\mathrm{HSIC}}_{k,\ell}(\Zn) 
    &= 
    \frac{1}{\big|\textbf{i}_4^N\big|} 
    \sum_{(i,j,r,s)\in \textbf{i}_4^N}
    h_{k,\ell}^{\mathrm{HSIC}}(Z_i, Z_j, Z_r, Z_s)
\end{align}
for characteristic kernels $k$ on $\R^{d_x}$ and $\ell$ on $\R^{d_y}$ \citep{gretton2015simpler}, and where
for $z_a=(x_a,y_a)\in\R^{d_x}\times\R^{d_y}$, $a=1,\dots,4$, we let
\begin{equation}
    \label{h_hsic}
    h_{k,\ell}^{\mathrm{HSIC}}(z_1, z_2, z_3, z_4)
    \coloneqq
    \frac{1}{4}
    h_k^{\mathrm{MMD}}(x_1, x_2; x_3, x_4)
    h_\ell^{\mathrm{MMD}}(y_1, y_2; y_3, y_4)
    .
\end{equation}

\textbf{Goodness-of-fit testing.}
For this problem, we are given a model density $p$ on $\R^d$ and i.i.d.\ samples $\Zn \coloneqq (Z_i)_{1\leq i\leq N}$ drawn from a density $q$ on $\R^d$. 
The aim is again to test $\mathcal{H}_0:p=q$ against $\mathcal{H}_1:p\neq q$; that is, we want to know if the samples have been drawn from the model.
\citet{chwialkowski2016kernel} and \citet{liu2016kernelized} both construct a nonparametric goodness-of-fit test using the {\em Kernel Stein Discrepancy} (KSD).
A quadratic-time KSD estimator can be computed as the second-order one-sample $U$-statistic,
\begin{equation}
    \label{Uksd}
    \widehat{\mathrm{KSD}}_{p,k}^2(\Zn)
    \coloneqq 
    \frac{1}{\abss{\textbf{i}_2^N}} 
    \sum_{(i,j)\in \textbf{i}_2^N}
    h_{k,p}^{\mathrm{KSD}}(Z_i,Z_j),
\end{equation}
where the {\em Stein kernel} $h_{k,p}^{\mathrm{KSD}}\colon\R^d\times\R^d\to\R$ is defined as
\begin{equation}
\label{h_ksd}
\begin{aligned}
    h_{k,p}^{\mathrm{KSD}}(x,y) \coloneqq\ 
    &\p{\nabla\log p(x)^\top \nabla\log p(y)} k(x,y)
    + \nabla\log p(y)^\top \nabla_x k(x,y) \\
    &+ \nabla\log p(x)^\top \nabla_y k(x,y)
    + \sum_{i=1}^d \frac{\partial}{\partial x_i \partial y_i}\, k(x,y).
\end{aligned}
\end{equation}
In order to guarantee consistency of the Stein goodness-of-fit test \citep[Theorem 2.2]{chwialkowski2016kernel}, we assume that the kernel $k$ is $C_0$-universal \citep[Definition 4.1]{carmeli2010vector} and that 
\begin{equation}
    \label{eq:ksdassumptions}
    \mathbb{E}_q \!\Big[h_{k,p}^{\mathrm{KSD}}(z,z) \Big]<\infty    
    \qquad \qquad \text{and} \qquad \qquad 
    \mathbb{E}_q \!\Bigg[\left\|\nabla \log\p{\frac{p(z)}{q(z)}}\right\|_2^2\Bigg]<\infty.
\end{equation}

\textbf{Quantile estimation.}
Multiple strategies have been proposed to estimate the quantiles of test statistics under the null for these three tests.
We primarily focus on the wild bootstrap approach \citep{chwialkowski2014wild}, though our results also hold using a parametric bootstrap for the goodness-of-fit setting \citep{schrab2022ksd}.
In \Cref{sec:minimaxproof}, we show that the same uniform separation rates can be derived for HSIC quadratic-time tests using permutations instead of a wild bootstrap.

More details on MMD, HSIC, KSD, and on quantile estimation are provided in \Cref{sec:detailedbackground}.

\section{Incomplete $U$-statistics for MMD, HSIC and KSD}
\label{sec:incomplete}

As presented above, the quadratic-time statistics for the two-sample (MMD), independence (HSIC) and goodness-of-fit (KSD) problems can be rewritten as $U$-statistics with kernels $h_k^{\mathrm{MMD}}$, $h_{k,\ell}^{\mathrm{HSIC}}$ and $h_{k,p}^{\mathrm{KSD}}$, respectively.
The computational cost of tests based on these $U$-statistics grows quadratically with the sample size.
When working with very large sample sizes, as it is often the case in real-world uses of those tests, this quadratic cost can become very problematic, and faster alternative tests are better adapted to this `big data' setting.
Multiple linear-time kernel tests have been proposed in the three testing frameworks (see \Cref{sec:introduction} for details).
We construct computationally efficient variants of the aggregated kernel tests proposed by \citet{fromont2013two}, \citet{albert2019adaptive}, \citet{kim2020minimax}, and \citet{schrab2021mmd,schrab2022ksd} for the three settings, with the aim of retaining the significant power advantages of the aggregation procedure observed for quadratic-time tests.
To this end, we propose to replace the quadratic-time $U$-statistics presented in \Cref{Ummd,Uhsic,Uksd} with second-order incomplete $U$-statistics \citep{blom1976some,janson1984asymptotic,lee1990ustatistic},
\begin{align}
    \label{Uincmmd}
    \overline{\mathrm{MMD}}^2_k\big(\Xm,\Yn; \D_N \big) 
    &\coloneqq
    \frac{1}{\abss{\D_N }} 
    \sum_{(i,j)\in \D_N }
    h_k^{\mathrm{MMD}}(X_i, X_{j}; Y_{i}, Y_{j}), \\
    \label{Uinchsic}
    \overline{\mathrm{HSIC}}_{k,\ell}\big(\Zn; \D_{\lfloor N/2\rfloor} \big) 
    &\coloneqq
    \frac{1}{\big|\D_{\lfloor N/2\rfloor} \big|} 
    \sum_{(i,j)\in \D_{\lfloor N/2\rfloor} }
    h_{k,\ell}^{\mathrm{HSIC}}\left(Z_i, Z_j, Z_{i+\lfloor N/2\rfloor}, Z_{j+\lfloor N/2\rfloor}\right), \\
    \label{Uincksd}
    \overline{\mathrm{KSD}}_{p,k}^2\big(\Zn; \D_N \big)
    &\coloneqq 
    \frac{1}{\abss{\D_N }} 
    \sum_{(i,j)\in \D_N }
    h_{k,p}^{\mathrm{KSD}}(Z_i,Z_j),
\end{align}
where for the two-sample problem we let $N\coloneqq \min(m,n)$,
and where the {\em design} $\D_b$ is a subset of $\textbf{i}_2^b$ (the set of all $2$-tuples drawn without replacement from $\{1,\dots,b\}$).
Note that $\D_{\lfloor N/2\rfloor} \subseteq \textbf{i}_2^{N/2} \subset \textbf{i}_2^N$.
The design can be deterministic. 
For example, for the two-sample problem with equal even sample sizes $m=n=N$,
the deterministic design 
$\D_N  = \{(2a-1, 2a): a=1,\dots,N/2\}$
corresponds to the MMD linear-time estimator proposed by \citet[Lemma 14]{gretton2012kernel}.
For fixed design size, the elements of the design can also be chosen at random without replacement, in which case the estimators in \Cref{Uincmmd,Uinchsic,Uincksd} become random quantities given the data.
For generality purposes, the results presented in this paper hold for both deterministic and random (without replacement) design choices while we focus on the deterministic design in our experiments.
By fixing the design sizes in \Cref{Uincmmd,Uinchsic,Uincksd} to be, for example, 
\begin{equation}
\label{eq:D}
\abss{\D_N } = \abss{\D_{\lfloor N/2 \rfloor}} = c N
\end{equation}
for some small constant $c\in\N\setminus\{0\}$, we obtain incomplete $U$-statistics  which can be computed in linear time.
Note that by pairing the samples $Z_i\coloneqq(X_i,Y_i)$, $i=1,\dots,N$ for the MMD case and $\widetilde Z_i\coloneqq\left(Z_i,Z_{i+\lfloor N/2\rfloor}\right)$, $i=1,\dots,\lfloor N/2\rfloor$ for the HSIC case, we observe that all three incomplete $U$-statistics of second order have the same form, with only the kernel functions and the design differing. 
The motivation for defining the estimators in \Cref{Uincmmd,Uinchsic} as incomplete $U$-statistics of order 2 (rather than of higher order) derives from the reasoning of \citet[Section 6]{kim2020minimax} for permuted complete $U$-statistics for the two-sample and independence problems (see \Cref{sec:hsicincref}).

\section{Quantile and variance bounds for incomplete $U$-statistics}
\label{sec:bounds}

In this section, we derive upper quantile and variance bounds for a second-order incomplete degenerate $U$-statistic with a generic degenerate kernel $h$, for some design $\D\subseteq \mathbf{i}_2^N$, defined as
$$
    \overline{U}\big(\Zn; \D \big)
    \coloneqq
    \frac{1}{\abs{\D}}
    \sum_{(i,j)\in\D }
    h(Z_i,Z_j).
$$
We will use these results to bound the quantiles and variances of our three test statistics for our hypothesis tests in \Cref{sec:rates}.
The derived bounds are of independent interest.

In the following lemma, building on the results of \cite{lee1990ustatistic}, we directly derive an upper bound on the variance of the incomplete $U$-statistic in terms of the sample size $N$ and of the design size $\abs{\D}$.

\begin{lemma}
    \label{lem:varbound}
    The variance of the incomplete $U$-statistic can be upper bounded in terms of the quantities 
    $
    \sigma_1^2 \coloneqq \mathrm{var}\!\p{\mathbb{E}\!\left[h(Z,Z')\big| Z'\right]}
    $
    and
    $
    \sigma_2^2 \coloneqq \mathrm{var}\!\p{h(Z,Z')}
    $ 
    with different bounds depending on the design choice.
    For deterministic (LHS) or random (RHS) design $\D$ and sample size $N$, we have 
    $$
      \mathrm{var}\!\left(\overline{U}\right)
      \lesssim
          \frac{N}{\abs{\D}}\sigma_1^2 + \frac{1}{\abs{\D}}\sigma_2^2
      \qquad\qquad \text{and} \qquad\qquad
      \mathrm{var}\!\left(\overline{U}\right)
      \lesssim 
          \frac{1}{N}\sigma_1^2 + \frac{1}{\abs{\D}} \sigma_2^2
      .
    $$
\end{lemma}
The proof of \Cref{lem:varbound} is deferred to \Cref{proof:varbound}.
We emphasize the fact that this variance bound also holds for random design with replacement, as considered by \citet{blom1976some} and \citet{lee1990ustatistic}.
For random design, we observe that if $\abs{\D} \asymp N^2$ then the bound is $\sigma_1^2 / N + \sigma_2^2 / N^2$ which is the variance bound of the complete $U$-statistic \citep[Lemma 10]{albert2019adaptive}. 
If $N \lesssim \abs{\D} \lesssim  N^2$, the variance bound is $\sigma_1^2 / N + \sigma_2^2 / \abs{\D}$, and if $\abs{\D} \lesssim N$ it is $\sigma_2^2 / \abs{\D}$ since $\sigma_1^2 \leq \sigma_2^2 / 2$ \citep[Equation 2.1]{blom1976some}.

\citet{kim2020minimax} develop exponential concentration bounds for permuted complete $U$-statistics, and \citet{clemencon2013maximal} study the uniform approximation of $U$-statistics by incomplete $U$-statistics. 
To the best of our knowledge, no quantile bounds have yet been obtained for incomplete $U$-statistics in the literature. 
While permutations are well-suited for complete $U$-statistics \citep{kim2020minimax}, using them with incomplete $U$-statistics results in having to compute new kernel values, which comes at an additional computational cost we would like to avoid. 
Restricting the set of permutations to those for which the kernel values have already been computed for the original incomplete $U$-statistic corresponds exactly to using a wild bootstrap \citep[Appendix B]{schrab2021mmd}.
Hence, we consider the wild bootstrapped second-order incomplete $U$-statistic
\begin{equation}
    \label{eq:Uwild}
    \overline{U}^{\epsilon}\big(\Zn; \D \big) 
    \coloneqq
    \frac{1}{\abs{\D}} 
    \sum_{(i,j)\in \D }
    \epsilon_i
    \epsilon_j
    h(Z_i,Z_j)
\end{equation}
for i.i.d.\ Rademacher random variables $\epsilon_1,\dots,\epsilon_N$ with values in $\{-1,1\}$, for which we derive an exponential concentration bound (quantile bound).
We note the in-depth work of \citet{chwialkowski2014wild} on the wild bootstrap procedure for kernel tests with applications to quadratic-time MMD and HSIC tests. We now provide exponential tail bounds for wild bootstrapped incomplete $U$-statistics.
\begin{lemma}
    \label{lem:Uincbound}
    There exists some constant $C>0$ such that, for every $t\geq 0$, we have
    \begin{align*}
        \PP_\epsilon\!\left(\abss{\overline{U}^{\epsilon}} \geq t \,\big|\, \Zn, \D\right)
        &~\leq~
        2\exp\!\p{
        -C\frac{t}{A_{\mathrm{inc}}}
        }
        ~\leq~
        2\exp\!\p{
        -C\frac{t}{A}
        }
    \end{align*}
    where
    $
        A_{\mathrm{inc}}^2 
        \coloneqq 
        \abs{\D}^{-2}
        \sum_{(i,j)\in \D }
        h(Z_i,Z_j)^2 
    $
    and
    $
        A^2 
        \coloneqq 
        \abs{\D}^{-2}
        \sum_{(i,j)\in \mathbf{i}_{2}^{N}}
        h(Z_i,Z_j)^2 
    $.
\end{lemma}
\Cref{lem:Uincbound} is proved in \Cref{proof:Uincbound}.
While the second bound in \Cref{lem:Uincbound} is less tight, it has the benefit of not depending on the choice of design $\D$ but only on its size $\abs{\D}$ which is usually fixed.

\section{Efficient kernel tests using incomplete $U$-statistics}
\label{sec:rates}

We now formally define the hypothesis tests obtained using the incomplete $U$-statistics with a wild bootstrap.
This is done for fixed kernel bandwidths $\lambda\in(0,\infty)^{d_x},\mu\in(0,\infty)^{d_y}$, for the kernels\footnote{Our results are presented for bandwidth selection, but they hold in the more general setting of kernel selection, as considered by \citet{schrab2022ksd}.
The goodness-of-fit results hold for a wider range of kernels including the IMQ (inverse multiquadric) kernel \citep{gorham2017measuring}, as in \citet{schrab2022ksd}.}
\begin{equation}
    \label{kernels}
    k_\lambda(x,y) \coloneqq \prod_{i=1}^{d_x} \frac{1}{\lambda_i}K_i\pp{\frac{x_i-y_i}{\lambda_i}},
    \qquad \qquad
    \ell_\mu(x,y) \coloneqq \prod_{i=1}^{d_y} \frac{1}{\mu_i}L_i\pp{\frac{x_i-y_i}{\mu_i}},
\end{equation}
for characteristic kernels $(x,y)\mapsto K_i(x-y)$, $(x,y)\mapsto L_i(x-y)$ on $\R\times\R$ for functions $K_i,L_i\in L^1(\R)\cap L^2(\R)$ integrating to 1.
We unify the notation for the three testing frameworks.
For the two-sample and goodness-of-fit problems, we work only with $k_\lambda$ and have $d= d_x$.
For the independence problem, we work with the two kernels $k_\lambda$ and $\ell_\mu$, and for ease of notation we let $d\coloneqq d_x+d_y$ and $\lambda_{d_x+i} \coloneqq \mu_i$ for $i=1,\dots,d_y$. We also simply write $p\coloneqq p_{xy}$ and $q\coloneqq p_x\otimes p_y$.
We let $\overline{U}_\lambda$ and $h_\lambda$ denote either 
$\overline{\mathrm{MMD}}^2_{k_\lambda}$ and $h_{k_\lambda}^{\mathrm{MMD}}$, 
or $\overline{\mathrm{HSIC}}_{k_\lambda,\ell_\mu}$ and $h_{k_\lambda,\ell_\mu}^{\mathrm{HSIC}}$, 
or $\overline{\mathrm{KSD}}_{p,k_\lambda}^2$ and $h_{k_\lambda,p}^{\mathrm{KSD}}$, respectively.
We denote the design size of the incomplete $U$-statistics in \Cref{Uincmmd,Uinchsic,Uincksd} by
$$
L\coloneqq \abss{\D_N} =  \abs{\D_{\lfloor N/2\rfloor}}.
$$
For the three testing frameworks, we estimate the quantiles of the test statistics by simulating the null hypothesis using a wild bootstrap, as done in the case of complete $U$-statistics by \citet{fromont2012kernels} and \citet{schrab2021mmd} for the two-sample problem, and by \citet{schrab2022ksd} for the goodness-of-fit problem.
This is done by considering the original test statistic $U^{B_1+1}_\lambda \coloneqq \overline{U}_\lambda$ together with $B_1$ wild bootstrapped incomplete $U$-statistics $U_\lambda^1,\dots,U_\lambda^{B_1}$ computed as in \Cref{eq:Uwild}, and estimating the $(1\!-\!\alpha)$-quantile with a Monte Carlo approximation
\begin{equation}
    \label{eq:bootstrapquantile}
    \widehat{q}_{1-\alpha}^{\,\lambda}
    \coloneqq\ \inf\!\bigg\{t \in \R: 1 - \alpha \leq \frac{1}{B_1+1}\sum_{b=1}^{B_1+1} \one\pp{ U_\lambda^b\leq t}\!\bigg\}
    = U_\lambda^{\bullet\lceil B_1(1-\alpha)\rceil},
\end{equation}
where $U_\lambda^{\bullet 1}\leq\cdots\leq U_\lambda^{\bullet B_1+1}$ are the sorted elements $U_\lambda^1,\dots,U_\lambda^{B_1+1}$.
The test $\Delta_\alpha^\lambda$ is defined as rejecting the null if the original test statistic $\overline{U}_\lambda$ is greater than the estimated $(1\!-\!\alpha)$-quantile, that is,
$$
\Delta_\alpha^\lambda(\Zn)\coloneqq \one\pp{\overline{U}_\lambda(\Zn)>\widehat{q}_{1-\alpha}^{\,\lambda}}.
$$ 
The resulting test has time complexity $\mathcal{O}\!\p{B_1L}$ where $L$ is the design size ($1\leq L \leq N(N-1)$).
We show in \Cref{prop:level} that the test $\Delta_\alpha^\lambda$ has well-calibrated asymptotic level for goodness-of-fit testing, and well-calibrated non-asymptotic level for two-sample and independence testing. The proof of the latter non-asymptotic guarantee is based on the exchangeability of $U_\lambda^1,\dots,U_\lambda^{B_1+1}$ under the null hypothesis along with the result of \citet[Lemma 1]{romano2005exact}. A similar proof strategy can be found in \citet[Proposition 2]{fromont2012kernels}, \citet[Proposition 1]{albert2019adaptive}, and \citet[Proposition 1]{schrab2021mmd}.
The exchangeability of wild bootstrapped incomplete $U$-statistics for independence testing does not follow directly from the mentioned works. 
We show this through the interesting connection between $h_{k,\ell}^{\mathrm{HSIC}}$ and $\{h_k^{\mathrm{MMD}}, h_\ell^{\mathrm{MMD}}\}$, the proof is deferred to \Cref{proof:level}.
\begin{proposition}
  \label{prop:level}
  The test $\Delta_\alpha^\lambda$ has level $\alpha\in(0,1)$, \emph{i.e.}
  $
  \PP_{\mathcal{H}_0}\!\pp{\Delta_\alpha^\lambda(\Zn)=1} \leq \alpha
  $.
  This holds non-asymptotically for the two-sample and independence cases, and asymptotically for  goodness-of-fit.\footnote{Level is non-asymptotic for the goodness-of-fit case using a parametric bootstrap \citep{schrab2022ksd}. For the goodness-of-fit setting, we also recall that the further assumptions in \Cref{eq:ksdassumptions} need to be satisfied.}
\end{proposition}
Having established the validity of the test $\Delta_\alpha^\lambda$, we now study power guarantees for it in terms of the $L^2$-norm of the difference in densities $\|p-q\|_2$. 
In \Cref{theo:fixed}, we show for the three tests that, if $\|p-q\|_2$ exceeds some threshold, we can guarantee high test power.
For the two-sample and independence problems, we derive uniform separation rates \citep{baraud2002non} over Sobolev balls
\begin{equation}
    \label{sobolev}
    \mathcal{S}_d^s(R) 
    \coloneqq 
    \Big\{ 
    f\in L^1\!\big(\R^d\big)\cap L^2\!\big(\R^d\big): \int_{\R^d} \|\xi\|^{2s}_2 |\widehat f(\xi)|^2 \mathrm{d} \xi \leq (2\pi)^d R^2
    \Big\},
\end{equation}
with radius $R>0$ and smoothness parameter $s>0$, where $\widehat f$ denotes the Fourier transform of $f$.
The uniform separation rate over $\mathcal{S}_d^s(R)$ is the smallest value of $t$ such that, for any alternative with $\|p-q\|_2>t$ and\footnote{We stress that we only assume $p-q\in\mathcal{S}_d^s(R)$ and not $p,q\in\mathcal{S}_d^s(R)$ as considered by \citet{li2019optimality}. Viewing $q$ as a perturbed version of $p$, we only require that the perturbation is smooth (\emph{i.e.} lies in a Sobolev ball).} $p-q\in\mathcal{S}_d^s(R)$, the probability of type II error of $\Delta_\alpha^\lambda$ can be controlled by $\beta\in(0,1)$.
Before presenting \Cref{theo:fixed}, we introduce further notation unified over the three testing frameworks; we define the integral transform $T_\lambda$ as 
\begin{equation}
\label{eq:integraltransform}
(T_\lambda f)(x) \coloneqq \int_{\R^d} f(x) \mathcal{K}_\lambda(x,y) \,\mathrm{d} y
\end{equation}
for $f\in L^2(\R^d)$, $x\in\R^d$, where 
$\mathcal{K}_\lambda\coloneqq k_\lambda$ for the two-sample problem, 
$\mathcal{K}_\lambda\coloneqq k_\lambda \otimes \ell_\mu$ for the independence problem, and
$\mathcal{K}_\lambda\coloneqq h_{k_\lambda,p}^{\mathrm{KSD}}$ for the goodness-of-fit problem.
Note that, for the two-sample and independence testing frameworks, since $\mathcal{K}_\lambda$ is translation-invariant, the integral transform corresponds to a convolution. However, this is not true for the goodness-of-fit setting as $h_{k_\lambda,p}^{\mathrm{KSD}}$ is not translation-invariant.
We are now in a position to present our main contribution in \Cref{theo:fixed}: we derive power guarantee conditions for our tests using incomplete $U$-statistics, and uniform separation rates over Sobolev balls for the two-sample and independence settings.
\begin{theorem}
    \label{theo:fixed}
    Suppose that the assumptions in \Cref{assump:fixed} hold, and consider $\lambda\in(0,\infty)^d$.

    (i) For sample size $N$ and design size $L$, if there exists some $C>0$ such that
    $$ 
      \|{p-q}\|^2_2 ~\geq~ \|{(p-q)-T_\lambda(p-q)}\|^2_2 + C \frac{N}{L}\frac{\ln\!\p{{1}/{\alpha}}}{\beta} \sigma_{2,\lambda},
    $$
    then 
    $\PP_{\mathcal{H}_1}\!\pp{\Delta_\alpha^\lambda(\Zn)=0}\leq \beta$ (type II error),
    where
    $\sigma_{2,\lambda}\lesssim 1/\sqrt{\LL}$ 
    for MMD and HSIC.

    (ii) Fix $R>0$ and $s>0$, and consider the bandwidths $\lambda_i^* \coloneqq (N/L)^{2/(4s+d)}$ for $i=1,\dots,d$. 
    For MMD and HSIC, the uniform separation rate of $\Delta_\alpha^{\lambda^*}$ over the Sobolev ball $\mathcal{S}_d^s(R)$ is (up to a constant)
    $$
        \pbig{L/N}^{-2s/(4s+d)}.
    $$
\end{theorem}
The proof of \Cref{theo:fixed} relies on the variance and quantile bounds presented in \Cref{lem:varbound,lem:Uincbound}, and also uses results of \citet{albert2019adaptive} and \citet{schrab2021mmd,schrab2022ksd} on complete $U$-statistics. 
The details can be found in \Cref{proof:fixed}.
The power condition in \Cref{theo:fixed} (i) corresponds to a variance-bias decomposition; for large bandwidths the bias term (first term) dominates, while for small bandwidths the variance term (second term which also controls the quantile) dominates.
While the power guarantees of \Cref{theo:fixed} hold for any design (either deterministic or uniformly random without replacement) of fixed size $L$, the choice of design still influences the performance of the test in practice.
The variance (but not its upper bound) depends on the choice of design; certain choices lead to minimum variance of the incomplete $U$-statistic \citep[Section 4.3.2]{lee1990ustatistic}.

The minimax (\emph{i.e.}~optimal) rate over the Sobolev ball $\Sb$ is $N^{-2s/(4s+d)}$ for the two-sample \citep[Theorem 5 (ii)]{li2019optimality} and independence (\citealp[Theorem 4]{albert2019adaptive}; \citealp[Corollary 5]{berrett2021optimal}) problems. 
The rate for our incomplete $U$-statistic test with time complexity $\mathcal{O}\!\p{B_1L}$ has the same dependence in the exponent as the minimax rate; %that is
$
\p{L/N}^{-2s/(4s+d)} = N^{-2s/(4s+d)} \p{N^2/L}^{2s/(4s+d)}
$
where $L\lesssim N^2$ 
with $L$ the design size and $N$ the sample size.
%with $L,N$ the design/sample sizes.% and $N$ the sample size.

\begin{tcolorbox}[colframe=black!25, colback=black!4] \vspace{-4pt}
\begin{center}
\end{center}
\vspace{-9pt}
\noindent
\vspace{3pt}
$\hspace{-0.1cm}\bullet$ If $L \asymp N^2$ then the test runs in quadratic time and we recover exactly the minimax rate.
\vspace{1pt}

$\hspace{-0.012cm}\bullet$ If $N \lesssim L \lesssim N^2$ then the rate still converges to 0; there is a trade-off between the cost 
\newline 
${\color{black!4}.}$ \hspace{-0.03cm} 
$(N^2/L)^{2s/(4s+d)}$ incurred in the minimax rate and the computational efficiency $\mathcal{O}\!\p{B_1L}$.

\vspace{4pt}
            $\hspace{-0.012cm}\bullet$ If $L\lesssim N$ then there is no guarantee that the rate converges to 0.
        \vspace{-5pt}
\end{tcolorbox}

To summarize, the tests we propose have computational cost $\mathcal{O}\!\p{B_1L}$ which can be specified by the user with the choice of the number of wild bootstraps $B_1$, and of the design size $L$ (as a function of the sample size $N$).
There is a trade-off between test power and computational cost.
We provide theoretical rates in terms of $L$ and $N$, working up to a constant. 
The rate is minimax optimal in the case where $L$ grows quadratically with $N$. 
We quantify exactly how, as the computational cost decreases  from quadratic to linear in the sample size, the rate deteriorates gradually from being minimax optimal to not being guaranteed to convergence to zero.
In our experiments, we use a design size which grows linearly with the sample size in order to compare our tests against other linear-time tests in the literature.
The assumption guaranteeing that the rate converges to $0$ is not satisfied in this setting, however, it would be satisfied for any faster growth of the design size (\emph{e.g.} $L\asymp N\log \log N$).

\section{Efficient aggregated kernel tests using incomplete $U$-statistics}
\label{sec:aggregation}
We now introduce our aggregated tests that combine single tests with different bandwidths. Our aggregation scheme is similar to those of \citet{fromont2013two}, \citet{albert2019adaptive} and \citet{schrab2021mmd,schrab2022ksd}, and can yield a test which is adaptive to the unknown smoothness parameter $s$ of the Sobolev ball $\Sb$, with relatively low price. Let $\Lambda$ be a finite collection of bandwidths, $(w_\lambda)_{\lambda\in\Lambda}$ be associated weights satisfying $\sum_{\lambda\in\Lambda}w_\lambda \leq 1$, and $u_\alpha$ be some correction term defined shortly in \Cref{eq:correction}. 
Then, using the incomplete $U$-statistic $\overline{U}_\lambda$, we define our aggregated test $\Delta_\alpha^\Lambda$ as 
$$
\Delta_\alpha^\Lambda(\Zn) 
\coloneqq 
\one\pBig{\overline{U}_\lambda(\Zn)>\widehat{q}_{1-u_\alpha w_\lambda}^{\,\lambda} \textrm{ for some } \lambda\in\Lambda}.
$$ 
The levels of the single tests are weighted and adjusted with a correction term
\begin{equation}  \label{eq:correction}
u_\alpha \coloneqq
\mathrm{sup}_{B_3}\!\left\{
    u\in\p{0,\,\min_{\lambda\in\Lambda} w_\lambda^{-1}} :
    \frac{1}{B_2}\sum_{b=1}^{B_2} \one\pp{
        \max_{\lambda\in\Lambda}\p{
            \widetilde U_\lambda^b - U_\lambda^{\bullet\lceil B_1(1-u w_\lambda)\rceil}
        } > 0
    }
    \leq \alpha
\right\},
\end{equation}
where the wild bootstrapped incomplete $U$-statistics $\widetilde U_\lambda^1,\dots, \widetilde U_\lambda^{B_2}$ computed as in \Cref{eq:Uwild} are used to perform a Monte Carlo approximation of the probability under the null, and where the supremum is estimated using $B_3$ steps of bisection method.
\Cref{prop:level}, along with the reasoning of \citet[Proposition 8]{schrab2021mmd}, ensures that $\Delta_\alpha^\Lambda$ has non-asymptotic level $\alpha$ for the two-sample and independence cases, and asymptotic level $\alpha$ for the goodness-of-fit case. 
We refer to the three aggregated tests constructed using incomplete $U$-statistics as MMDAggInc, HSICAggInc and KSDAggInc.
The computational complexity of those tests is $\mathcal{O}\!\p{\abs{\Lambda}(B_1+B_2)L}$, which means, for example, that if $L \asymp N$ as in \Cref{eq:D}, the tests run efficiently in linear time in the sample size. 

We formally record error guarantees of $\Delta^\Lambda_\alpha$ and derive uniform separation rates over Sobolev balls.
\begin{theorem}
    \label{theo:agg}
    Suppose that the assumptions in \Cref{assump:agg} hold, and consider a collection $\Lambda$.
    
    (i) For sample size $N$ and design size $L$, if there exists some $C>0$ such that
    $$ 
      \|{p-q}\|^2_2 ~\geq~ \min_{\lambda\in\Lambda}\p{\|{(p-q)-T_\lambda(p-q)}\|^2_2 + C \frac{N}{L}\frac{\ln\!\pbig{{1}/{(\alpha w_\lambda)}}}{\beta}\sigma_{2,\lambda}},
    $$
    then
    $\PP_{\mathcal{H}_1}\!\pp{\Delta_\alpha^\Lambda(\Zn)=0}\leq \beta$ (type II error),
    where
    $\sigma_{2,\lambda}\lesssim 1 /\sqrt{\LL}$ 
    for MMD and HSIC.

    (ii) Assume $L>N$ so that $\ln(\ln(L/N))$ is well-defined.
    Consider the collections of bandwidths and weights (independent of the parameters $s$ and $R$ of the Sobolev ball $\mathcal{S}_d^s(R)$)
    $$
        \Lambda \coloneqq \Big\{\big(2^{-\ell},\dots,2^{-\ell}\big) \in (0,\infty)^d: \ell \in \Big\{1,\dots, \Big\lceil\frac{2}{d}\log_2\!\Big(\frac{L/N}{\ln(\ln(L/N))}\Big)\Big\rceil\Big\}\Big\},
        \quad
        w_\lambda \coloneqq \frac{6}{\pi^{2}\ell^{2}}.
    $$
    For the two-sample and independence problems, the uniform separation rate of $\Delta_\alpha^\Lambda$ over the Sobolev balls $\big\{\mathcal{S}_d^s(R):R>0,s>0\big\}$ is (up to a constant)
    $$
        \p{\frac{L/N}{\ln\pp{\ln\pp{L/N}}}}^{-2s/(4s+d)}.
    $$
\end{theorem}
The extension from \Cref{theo:fixed} to \Cref{theo:agg} has been proved for complete $U$-statistics in the two-sample \citep{fromont2013two,schrab2021mmd}, independence \citep{albert2019adaptive} and goodness-of-fit \citep{schrab2022ksd} testing frameworks. 
The proof of \Cref{theo:agg} follows with the same reasoning by simply replacing $N$ with $L/N$ as we work with incomplete $U$-statistics; this `replacement' is theoretically justified by \Cref{theo:fixed}.
\Cref{theo:agg} shows that the aggregated test $\Delta_\alpha^\Lambda$ is {\em adaptive} over  Sobolev balls $\big\{\mathcal{S}_d^s(R):R>0,s>0\big\}$: the test $\Delta_\alpha^\Lambda$ does not depend on the unknown smoothness parameter $s$ (unlike $\Delta_\alpha^{\lambda^*}$ in \Cref{theo:fixed}) and achieves the minimax rate, up to an iterated logarithmic factor, and up to the cost incurred for efficiency of the test (\emph{i.e.}~$L/N$ instead of $N$).

\section{Minimax optimal permuted quadratic-time aggregated independence test}
\label{sec:minimaxproof}

Considering \Cref{theo:agg} with our incomplete $U$-statistic with full design $\D=\mathbf{i}_2^N$ for which $L \asymp N^2$, we have proved that the quadratic-time two-sample and independence aggregated tests using a wild bootstrap achieve the rate
$
\p{N/{\ln\pp{\ln\pp{N}}}}^{-2s/(4s+d)}
$
over the Sobolev balls $\big\{\mathcal{S}_d^s(R):R>0,s>0\big\}$.
This is the minimax rate \citep{li2019optimality,albert2019adaptive}, up to some iterated logarithmic term.
For the two-sample problem, \citet{kim2020minimax} and \citet{schrab2021mmd} show that this optimality result also holds when using complete $U$-statistics with permutations.
Whether the equivalent statement for the independence test with permutations holds has not yet been addressed; the rate can be proved using theoretical (unknown) quantiles with a Gaussian kernel \citep{albert2019adaptive}, but has not yet been proved using permutations.
\citet[Proposition 8.7]{kim2020minimax} consider this problem, again using a Gaussian kernel, but they do not obtain the correct dependence on $\alpha$ (\emph{i.e.} they obtain $\alpha^{-1/2}$ rather than $\ln\pp{1/\alpha}$), hence they cannot recover the desired rate.
As pointed out by \citet[Section 8]{kim2020minimax}: `It remains an open question as to whether [the power guarantee] continues to hold when $\alpha^{-1/2}$ is replaced by $\ln\pp{1/\alpha}$'.
We now prove that we can improve the $\alpha$-dependence to  
$\ln\pp{1/\alpha}^{3/2}$ for any bounded kernel of the form of \Cref{kernels}, and that this allows us to obtain the desired rate over Sobolev balls $\big\{\mathcal{S}_d^s(R):R>0,s>d/4\big\}$.
The assumption $s>d/4$ imposes a stronger smoothness restriction on $p-q\in\Sb$, which is similarly also considered by \citet{li2019optimality}.
\begin{theorem}
    \label{theo:indminimax}
    Consider the quadratic-time independence test using the complete $U$-statistic HSIC estimator with a quantile estimated using permutations as done by \citet[Proposition 8.7]{kim2020minimax}, with kernels as in \Cref{kernels} for bounded functions $K_i$ and $L_j$ for $i=1,\dots,d_x$, $j=1,\dots,d_y$.

    (i) Suppose that the assumptions in \Cref{assump:fixed} hold.
    For fixed $R>0$, $s>d/4$, and bandwidths $\lambda_i^* \coloneqq N^{-2/(4s+d)}$ for $i=1,\dots,d$, the probability of type II error of the test is controlled by $\beta$ when
    $$ 
      \|{p-q}\|^2_2 ~\geq~ \|{(p-q)-T_{\lambda^*}(p-q)}\|^2_2 + C \frac{1}{N}\frac{\ln\!\p{{1}/{\alpha}}^{3/2}}{\beta \sqrt{\lambda_1^*\cdots\lambda_d^*}}
      \quad
      \text{for some constant $C>0$}.
    $$
    The uniform separation rate over the Sobolev ball $\mathcal{S}_d^s(R)$ is, up to a constant,
    $
        N^{-2s/(4s+d)}
    $.

    (ii) 
    Suppose that the assumptions in \Cref{assump:agg} hold.
    The uniform separation rate over the Sobolev balls $\big\{\mathcal{S}_d^s(R):R>0, s>d/4\big\}$ is
    $
        \pbig{{N}/\ln\pp{\ln\pp{N}}\!}^{-2s/(4s+d)}
    $, up to a constant,
    with the collections
    $$
        \Lambda \coloneqq \Big\{\big(2^{-\ell},\dots,2^{-\ell}\big) \in (0,\infty)^d: \ell \in \Big\{1,\dots, \Big\lceil\frac{2}{d}\log_2\!\Big(\frac{N}{\ln(\ln(N))}\Big)\Big\rceil\Big\}\Big\},
        \quad
        w_\lambda \coloneqq \frac{6}{\pi^{2}\ell^{2}}.
    $$
\end{theorem}
The proof of \Cref{theo:indminimax}, in \Cref{proof:indminimax},  uses the exponential concentration bound of \citet[Theorem 6.3]{kim2020minimax} for permuted complete $U$-statistics.
Another possible approach to obtain the correct dependency on $\alpha$ is to employ the sample-splitting method proposed by \citet[Section 8.3]{kim2020minimax} in order to transform the independence problem into a two-sample problem.
While this indirect approach leads to a logarithmic factor in $\alpha$, the practical power would be suboptimal due to an inefficient use of the data from sample splitting. 
\Cref{theo:indminimax} (i) shows that a $\ln\!\p{{1}/{\alpha}}^{3/2}$ dependence is achieved by the more practical permutation-based HSIC test.
\Cref{theo:indminimax} (ii) demonstrates that this leads to a minimax optimal rate for the aggregated HSIC test, up to the $\ln(\ln(N))$ cost for adaptivity.

\section{Experiments}
\label{sec:experiments}

For the two-sample problem, we consider testing samples drawn from a uniform density on $[0,1]^d$ against samples drawn from a perturbed uniform density.
For the independence problem, the joint density is a perturbed uniform density on $[0,1]^{d_x+d_y}$, the marginals are then simply uniform densities.
Those perturbed uniform densities can be shown to lie in Sobolev balls \citep{li2019optimality,albert2019adaptive}, to which our tests are adaptive.
For the goodness-of-fit problem, we use a Gaussian-Bernoulli Restricted Boltzmann Machine as first considered by \citet{liu2016kernelized} in this testing framework.
We use collections of 21 bandwidths for MMD and HSIC and of $25$ bandwidth pairs for HSIC; more details on the experiments (\emph{e.g.} model and test parameters) are presented in \Cref{sec:detailedexperiments}.

We consider our incomplete aggregated tests MMDAggInc, HSICAggInc and KSDAggInc, with parameter $R\in \{1,\dots,N-1\}$ which fixes the deterministic design to consist of the first $R$ sub-diagonals of the $N\times N$ matrix, \emph{i.e.}
$\D \coloneqq \{(i,i+r): i = 1, \dots, N-r \text{ for } r = 1, \dots, R\}$ with size
$\abs{\D} = RN - R(R-1)/2$.
We run our incomplete tests with $R\in\{1, 100, 200\}$ and also the complete test using the full design $\D=\mathbf{i}_2^N$.
We compare their performances with current linear-time state-of-the-art tests:
ME, SCF, FSIC and FSSD \citep{jitkrittum2016interpretable,jitkrittum2017adaptive,jitkrittum2017linear} which evaluate the witness functions at a finite set of locations chosen to maximize the power, 
Cauchy RFF (random Fourier feature) and L1 IMQ \citep{huggins2018random} which are random feature Stein discrepancies,
LSD \citep{grathwohl2020learning} which uses a neural network to learn the Stein discrepancy,
and OST PSI \citep{kubler2020learning} which performs kernel selection using post selection inference.

\begin{figure}
  \centering
  \includegraphics[width=\textwidth]{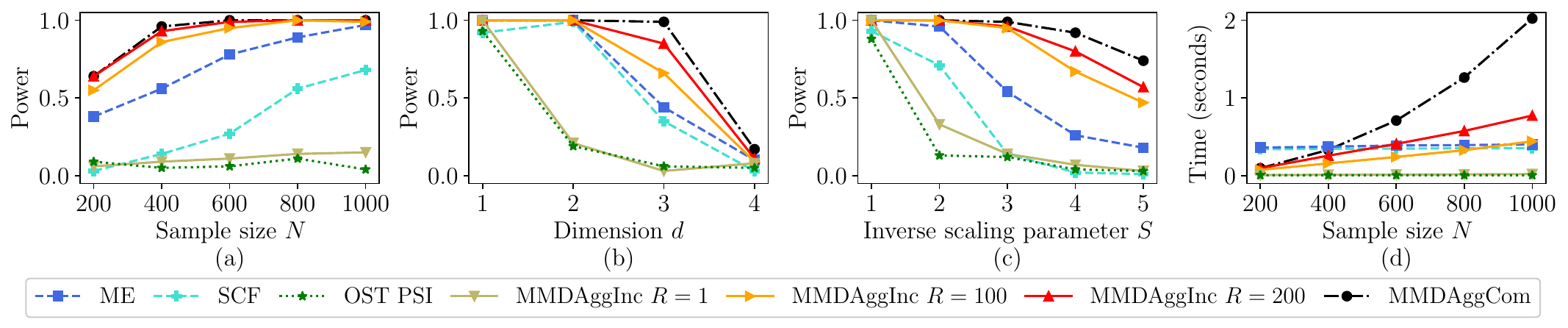}
  \includegraphics[width=\textwidth]{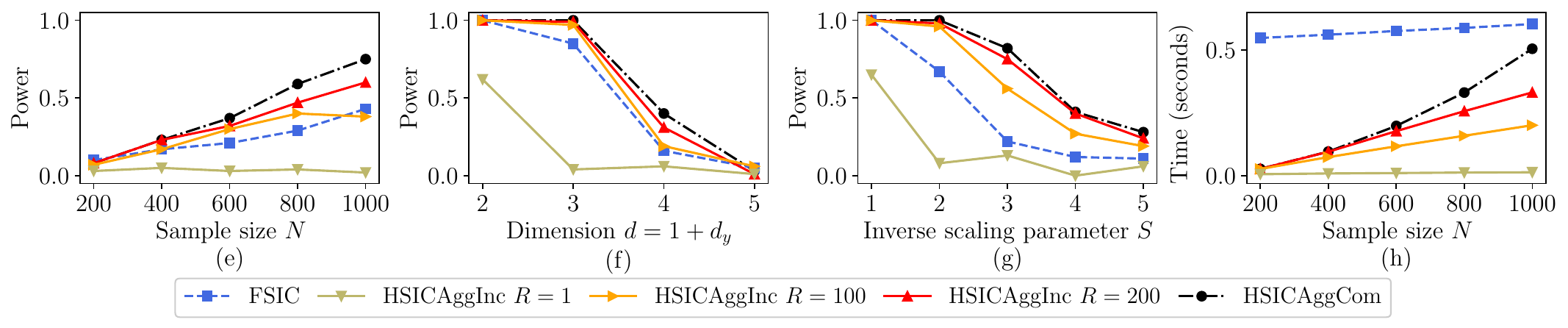}
  \includegraphics[width=\textwidth]{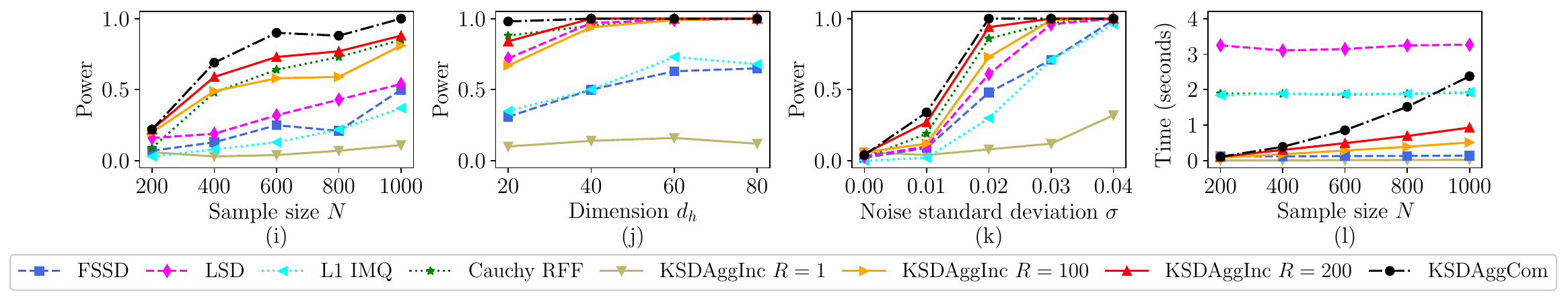}
  \caption{%
  Two-sample {\em (a--d)} and independence {\em (e--h)} experiments using perturbed uniform densities.
  Goodness-of-fit {\em (i--l)} experiment using a Gaussian-Bernoulli Restricted Boltzmann Machine.
  The power results are averaged over 100 repetitions and the runtimes over 20 repetitions. 
  \label{fig:experiments}
  }
\end{figure}

Similar trends are observed across all our experiments in \Cref{fig:experiments}, for the three testing frameworks, when varying the sample size, the dimension, and the difficulty of the problem (scale of perturbations or noise level).
The linear-time tests AggInc $R=200$ almost match the power obtained by the quadratic-time tests AggCom in all settings (except in \Cref{fig:experiments} (i) where the difference is larger) while being computationally much more efficient as can be seen in \Cref{fig:experiments} (d, h, l). 
The incomplete tests with $R=100$ have power only slightly below the ones using $R=200$, and run roughly twice as fast (\Cref{fig:experiments} (d, h, l)).
In all experiments, those three tests (AggInc $R=100, 200$ and AggCom) have significantly higher power than the linear-time tests which optimize test locations (ME, SCF, FSIC and FSSD); 
in the two-sample case the aggregated tests run faster for small sample size but slower for large sample size, 
in the independence case the aggregated tests run much faster,
and in the goodness-of-fit case FSSD runs faster.
While both types of tests are linear, we note that the runtimes of the tests of \citet{jitkrittum2016interpretable,jitkrittum2017adaptive,jitkrittum2017linear} increase slower with the sample size than those of our aggregated tests with $R=100,200$, but a fixed computational cost is incurred for their optimization step, even for small sample sizes.
In the goodness-of-fit framework, L1 IMQ performs similarly to FSSD which is in line with the results presented by \citet[Figure 4d]{huggins2018random} who consider the same experiment.
All other goodness-of-fit tests (except KSDAggInc $R=1$) achieve much higher test power.
Cauchy RFF and KSDAggInc $R=200$ obtain similar power in almost all the experiments.
While KSDAggInc $R=200$ runs much faster in the experiments presented\footnote{The runtimes in \Cref{fig:experiments} (d, h, l) can also vary due to the different implementations of the respective authors.}, it seems that the KSDAggInc runtimes increase more steeply with the sample size than the Cauchy RFF and L1 IMQ runtimes (see \Cref{sec:ndraws} for details).
LSD matches the power of KSDAggInc $R=100$ when varying the noise level in \Cref{fig:experiments} (k) (KSDAggInc $R=200$ has higher power), and when varying the hidden dimension in \Cref{fig:experiments} (j) where $d_x=100$.
When varying the sample size in \Cref{fig:experiments} (i), both KSDAggInc tests with $R=100,200$ achieve much higher power than LSD.
Unsurprisingly, AggInc $R=1$, which runs much faster than all the aforementioned tests, has low power in every experiment.
For the two-sample problem, it obtains slightly higher power than OST PSI which runs even faster.
We include more experiments in \Cref{sec:additional_experiments}: 
we present experiments on the MNIST dataset (same trends are observed),
we use different collection of bandwidths,
we verify that all tests have well-calibrated levels,
and illustrate the benefits of the aggregation procedure.

\section{Acknowledgements}

Antonin Schrab acknowledges support from the U.K.\ Research and Innovation (EP/S021566/1).
Ilmun Kim acknowledges support from the Yonsei University Research Fund of 2021-22-0332, and from the Basic Science Research Program through the National Research Foundation of Korea funded by the Ministry of Education (2022R1A4A1033384).
Benjamin Guedj acknowledges partial support by the U.S.\ Army Research Laboratory and the U.S.\ Army Research Office, and by the U.K.\ Ministry of Defence and the U.K.\ Engineering and Physical Sciences Research Council (EP/R013616/1), and by the French National Agency for Research (ANR-18-CE40-0016-01 \& ANR-18-CE23-0015-02).
Arthur Gretton acknowledges support from the Gatsby Charitable Foundation.

\clearpage
\bibliographystyle{apalike}
\bibliography{biblio_inc}

\begin{thebibliography}{}

\bibitem[Albert et~al., 2022]{albert2019adaptive}
Albert, M., Laurent, B., Marrel, A., and Meynaoui, A. (2022).
\newblock Adaptive test of independence based on {HSIC} measures.
\newblock {\em The Annals of Statistics}, 50(2):858--879.

\bibitem[Aronszajn, 1950]{azonszajn1950theory}
Aronszajn, N. (1950).
\newblock Theory of reproducing kernels.
\newblock {\em Transactions of the American Mathematical Society},
  68(3):337--404.

\bibitem[Balasubramanian et~al., 2021]{balasubramanian2021optimality}
Balasubramanian, K., Li, T., and Yuan, M. (2021).
\newblock On the optimality of kernel-embedding based goodness-of-fit tests.
\newblock {\em Journal of Machine Learning Research}, 22(1).

\bibitem[Baraud, 2002]{baraud2002non}
Baraud, Y. (2002).
\newblock Non-asymptotic minimax rates of testing in signal detection.
\newblock {\em Bernoulli}, 1(8(5):577--606).

\bibitem[Berrett et~al., 2021]{berrett2021optimal}
Berrett, T.~B., Kontoyiannis, I., and Samworth, R.~J. (2021).
\newblock Optimal rates for independence testing via u-statistic permutation
  tests.
\newblock {\em The Annals of Statistics}, 49(5):2457--2490.

\bibitem[Blom, 1976]{blom1976some}
Blom, G. (1976).
\newblock Some properties of incomplete {U}-statistics.
\newblock {\em Biometrika}, 63(3):573--580.

\bibitem[Carmeli et~al., 2010]{carmeli2010vector}
Carmeli, C., De~Vito, E., Toigo, A., and Umanit{\'a}, V. (2010).
\newblock {Vector valued reproducing kernel Hilbert spaces and universality}.
\newblock {\em Analysis and Applications}, 8(01):19--61.

\bibitem[Chebyshev, 1899]{chebyshev1899oeuvres}
Chebyshev, P.~L. (1899).
\newblock Oeuvres.
\newblock {\em Commissionaires de l'Acad{\'e}mie Imp{\'e}riale des Sciences},
  1.

\bibitem[Cherfaoui et~al., 2022]{cherfaoui2022discrete}
Cherfaoui, F., Kadri, H., Anthoine, S., and Ralaivola, L. (2022).
\newblock A discrete {RKHS} standpoint for {Nystr{\"o}m} {MMD}.
\newblock {\em HAL preprint hal-03651849}.

\bibitem[Chwialkowski et~al., 2014]{chwialkowski2014wild}
Chwialkowski, K., Sejdinovic, D., and Gretton, A. (2014).
\newblock A wild bootstrap for degenerate kernel tests.
\newblock In {\em Advances in neural information processing systems}, pages
  3608--3616.

\bibitem[Chwialkowski et~al., 2016]{chwialkowski2016kernel}
Chwialkowski, K., Strathmann, H., and Gretton, A. (2016).
\newblock A kernel test of goodness of fit.
\newblock In {\em International Conference on Machine Learning}, pages
  2606--2615. PMLR.

\bibitem[Chwialkowski et~al., 2015]{chwialkowski2015fast}
Chwialkowski, K.~P., Ramdas, A., Sejdinovic, D., and Gretton, A. (2015).
\newblock Fast two-sample testing with analytic representations of probability
  measures.
\newblock In {\em Advances in Neural Information Processing Systems},
  volume~28, pages 1981--1989.

\bibitem[Cl{\'e}men{\c{c}}on et~al., 2013]{clemencon2013maximal}
Cl{\'e}men{\c{c}}on, S., Robbiano, S., and Tressou, J. (2013).
\newblock Maximal deviations of incomplete {U}-statistics with applications to
  empirical risk sampling.
\newblock In {\em Proceedings of the 2013 SIAM International Conference on Data
  Mining}, pages 19--27. SIAM.

\bibitem[de~la Pe{\~n}a and Gin{\'e}, 1999]{pena1999decoupling}
de~la Pe{\~n}a, V.~H. and Gin{\'e}, E. (1999).
\newblock {\em Decoupling: From Dependence to Independence}.
\newblock Springer Science \& Business Media.

\bibitem[Dinh et~al., 2017]{dinh2016density}
Dinh, L., Sohl{-}Dickstein, J., and Bengio, S. (2017).
\newblock Density estimation using real {NVP}.
\newblock In {\em International Conference on Learning Representations}.

\bibitem[Duembgen, 1998]{dumbgen1998symmetrization}
Duembgen, L. (1998).
\newblock Symmetrization and decoupling of combinatorial random elements.
\newblock {\em Statistics \& probability letters}, 39(4):355--361.

\bibitem[Dvoretzky et~al., 1956]{dvoretzky1956asymptotic}
Dvoretzky, A., Kiefer, J., and Wolfowitz, J. (1956).
\newblock Asymptotic minimax character of the sample distribution function and
  of the classical multinomial estimator.
\newblock {\em The Annals of Mathematical Statistics}, pages 642--669.

\bibitem[Fithian et~al., 2014]{fithian2014optimal}
Fithian, W., Sun, D., and Taylor, J. (2014).
\newblock Optimal inference after model selection.
\newblock {\em arXiv preprint arXiv:1410.2597}.

\bibitem[Freidling et~al., 2021]{freidling2021post}
Freidling, T., Poignard, B., Climente{-}Gonz{\'{a}}lez, H., and Yamada, M.
  (2021).
\newblock {Post-selection inference with HSIC-Lasso}.
\newblock In {\em International Conference on Machine Learning}, pages
  3439--3448. {PMLR}.

\bibitem[Fromont et~al., 2012]{fromont2012kernels}
Fromont, M., Laurent, B., Lerasle, M., and Reynaud{-}Bouret, P. (2012).
\newblock Kernels based tests with non-asymptotic bootstrap approaches for
  two-sample problems.
\newblock In {\em {Conference on Learning Theory}}, PMLR.

\bibitem[Fromont et~al., 2013]{fromont2013two}
Fromont, M., Laurent, B., and Reynaud-Bouret, P. (2013).
\newblock The two-sample problem for {P}oisson processes: Adaptive tests with a
  nonasymptotic wild bootstrap approach.
\newblock {\em The Annals of Statistics}, 41(3):1431--1461.

\bibitem[Fukumizu et~al., 2008]{fukumizu2008kernel}
Fukumizu, K., Gretton, A., Sun, X., and Sch{\"o}lkopf, B. (2008).
\newblock Kernel measures of conditional dependence.
\newblock In {\em Advances in Neural Information Processing Systems}, volume~1,
  pages 489--496.

\bibitem[Gorham and Mackey, 2017]{gorham2017measuring}
Gorham, J. and Mackey, L. (2017).
\newblock Measuring sample quality with kernels.
\newblock In {\em International Conference on Machine Learning}, pages
  1292--1301. PMLR.

\bibitem[Grathwohl et~al., 2020]{grathwohl2020learning}
Grathwohl, W., Wang, K.-C., Jacobsen, J.-H., Duvenaud, D., and Zemel, R.
  (2020).
\newblock Learning the {Stein} discrepancy for training and evaluating
  energy-based models without sampling.
\newblock In {\em International Conference on Machine Learning}, pages
  3732--3747. PMLR.

\bibitem[Gretton, 2015]{gretton2015simpler}
Gretton, A. (2015).
\newblock A simpler condition for consistency of a kernel independence test.
\newblock {\em arXiv preprint arXiv:1501.06103}.

\bibitem[Gretton et~al., 2012a]{gretton2012kernel}
Gretton, A., Borgwardt, K.~M., Rasch, M.~J., Sch{\"o}lkopf, B., and Smola, A.
  (2012a).
\newblock A kernel two-sample test.
\newblock {\em Journal of Machine Learning Research}, 13:723--773.

\bibitem[Gretton et~al., 2009]{gretton2009fast}
Gretton, A., Fukumizu, K., Harchaoui, Z., and Sriperumbudur, B.~K. (2009).
\newblock A fast, consistent kernel two-sample test.
\newblock {\em Advances in Neural Information Processing Systems}, 22.

\bibitem[Gretton et~al., 2008]{gretton2008kernel}
Gretton, A., Fukumizu, K., Teo, C.~H., Song, L., Sch{\"o}lkopf, B., and Smola,
  A.~J. (2008).
\newblock A kernel statistical test of independence.
\newblock In {\em Advances in Neural Information Processing Systems}, volume~1,
  pages 585--592.

\bibitem[Gretton et~al., 2005]{gretton2005kernel}
Gretton, A., Herbrich, R., Smola, A., Bousquet, O., and Sch{\"o}lkopf, B.
  (2005).
\newblock Kernel methods for measuring independence.
\newblock {\em Journal of Machine Learning Research}, 6:2075--2129.

\bibitem[Gretton et~al., 2012b]{gretton2012optimal}
Gretton, A., Sejdinovic, D., Strathmann, H., Balakrishnan, S., Pontil, M.,
  Fukumizu, K., and Sriperumbudur, B.~K. (2012b).
\newblock Optimal kernel choice for large-scale two-sample tests.
\newblock In {\em Advances in Neural Information Processing Systems}, volume~1,
  pages 1205--1213.

\bibitem[Ho and Shieh, 2006]{ho2006two}
Ho, H.-C. and Shieh, G.~S. (2006).
\newblock Two-stage {U}-statistics for hypothesis testing.
\newblock {\em Scandinavian journal of statistics}, 33(4):861--873.

\bibitem[Hoeffding, 1992]{hoeffding1992class}
Hoeffding, W. (1992).
\newblock A class of statistics with asymptotically normal distribution.
\newblock In {\em Breakthroughs in Statistics}, pages 308--334. Springer.

\bibitem[Huggins and Mackey, 2018]{huggins2018random}
Huggins, J. and Mackey, L. (2018).
\newblock Random feature {S}tein discrepancies.
\newblock {\em Advances in Neural Information Processing Systems}, 31.

\bibitem[Janson, 1984]{janson1984asymptotic}
Janson, S. (1984).
\newblock The asymptotic distributions of incomplete {U}-statistics.
\newblock {\em Zeitschrift f{\"u}r Wahrscheinlichkeitstheorie und Verwandte
  Gebiete}, 66(4):495--505.

\bibitem[Jitkrittum et~al., 2016]{jitkrittum2016interpretable}
Jitkrittum, W., Szab{\'o}, Z., Chwialkowski, K.~P., and Gretton, A. (2016).
\newblock Interpretable distribution features with maximum testing power.
\newblock In {\em Advances in Neural Information Processing Systems},
  volume~29, pages 181--189.

\bibitem[Jitkrittum et~al., 2017a]{jitkrittum2017adaptive}
Jitkrittum, W., Szab{\'{o}}, Z., and Gretton, A. (2017a).
\newblock An adaptive test of independence with analytic kernel embeddings.
\newblock In {\em International Conference on Machine Learning (ICML)}, pages
  1742--1751.

\bibitem[Jitkrittum et~al., 2017b]{jitkrittum2017linear}
Jitkrittum, W., Xu, W., Szab{\'{o}}, Z., Fukumizu, K., and Gretton, A. (2017b).
\newblock A linear-time kernel goodness-of-fit test.
\newblock In {\em Advances in Neural Information Processing Systems}, pages
  262--271.

\bibitem[Key et~al., 2021]{key2021composite}
Key, O., Fernandez, T., Gretton, A., and Briol, F.-X. (2021).
\newblock Composite goodness-of-fit tests with kernels.
\newblock {\em arXiv preprint arXiv:2111.10275}.

\bibitem[Kim et~al., 2022]{kim2020minimax}
Kim, I., Balakrishnan, S., and Wasserman, L. (2022).
\newblock Minimax optimality of permutation tests.
\newblock {\em The Annals of Statistics}, 50(1):225--251.

\bibitem[Kingma and Ba, 2014]{kingma2014adam}
Kingma, D.~P. and Ba, J. (2014).
\newblock Adam: A method for stochastic optimization.
\newblock {\em arXiv preprint arXiv:1412.6980}.

\bibitem[Kingma and Dhariwal, 2018]{kingma2018glow}
Kingma, D.~P. and Dhariwal, P. (2018).
\newblock Glow: Generative flow with invertible 1x1 convolutions.
\newblock In {\em Advances in Neural Information Processing Systems}, pages
  10236--10245.

\bibitem[K{\"u}bler et~al., 2020]{kubler2020learning}
K{\"u}bler, J.~M., Jitkrittum, W., Sch{\"o}lkopf, B., and Muandet, K. (2020).
\newblock Learning kernel tests without data splitting.
\newblock In {\em Advances in Neural Information Processing Systems 33}, pages
  6245--6255. Curran Associates, Inc.

\bibitem[K{\"u}bler et~al., 2022]{kubler2022witness}
K{\"u}bler, J.~M., Jitkrittum, W., Sch{\"o}lkopf, B., and Muandet, K. (2022).
\newblock A witness two-sample test.
\newblock In {\em International Conference on Artificial Intelligence and
  Statistics}, pages 1403--1419. PMLR.

\bibitem[LeCun et~al., 2010]{lecun2010mnist}
LeCun, Y., Cortes, C., and Burges, C. (2010).
\newblock {MNIST handwritten digit database. AT\&T Labs}.

\bibitem[Lee, 1990]{lee1990ustatistic}
Lee, J. (1990).
\newblock {\em ${U}$-statistics: {T}heory and {P}ractice}.
\newblock Citeseer.

\bibitem[Lee et~al., 2016]{lee2016exact}
Lee, J.~D., Sun, D.~L., Sun, Y., and Taylor, J.~E. (2016).
\newblock {Exact post-selection inference, with application to the {L}asso}.
\newblock {\em The Annals of Statistics}, 44(3):907--927.

\bibitem[Leucht and Neumann, 2013]{leucht2013dependent}
Leucht, A. and Neumann, M.~H. (2013).
\newblock {Dependent wild bootstrap for degenerate U- and V-statistics}.
\newblock {\em Journal of Multivariate Analysis}, 117:257--280.

\bibitem[Li and Yuan, 2019]{li2019optimality}
Li, T. and Yuan, M. (2019).
\newblock On the optimality of gaussian kernel based nonparametric tests
  against smooth alternatives.
\newblock {\em arXiv preprint arXiv:1909.03302}.

\bibitem[Lim et~al., 2020]{lim2020more}
Lim, J.~N., Yamada, M., Jitkrittum, W., Terada, Y., Matsui, S., and Shimodaira,
  H. (2020).
\newblock More powerful selective kernel tests for feature selection.
\newblock In {\em International Conference on Artificial Intelligence and
  Statistics}, pages 820--830. {PMLR}.

\bibitem[Lim et~al., 2019]{lim2019kernel}
Lim, J.~N., Yamada, M., Sch{\"{o}}lkopf, B., and Jitkrittum, W. (2019).
\newblock Kernel {Stein} tests for multiple model comparison.
\newblock In {\em Advances in Neural Information Processing Systems}, pages
  2240--2250.

\bibitem[Liu et~al., 2016]{liu2016kernelized}
Liu, Q., Lee, J., and Jordan, M. (2016).
\newblock {A kernelized {Stein} discrepancy for goodness-of-fit tests}.
\newblock In {\em International Conference on Machine Learning}, pages
  276--284. PMLR.

\bibitem[Massart, 1990]{massart1990tight}
Massart, P. (1990).
\newblock {The tight constant in the Dvoretzky-Kiefer-Wolfowitz inequality}.
\newblock {\em The Annals of Probability}, 18(3):1269--1283.

\bibitem[M{\"u}ller, 1997]{muller1997integral}
M{\"u}ller, A. (1997).
\newblock Integral probability metrics and their generating classes of
  functions.
\newblock {\em Advances in Applied Probability}, 1:429--443.

\bibitem[Rahimi and Recht, 2007]{rahimi2007random}
Rahimi, A. and Recht, B. (2007).
\newblock Random features for large-scale kernel machines.
\newblock In {\em Advances in Neural Information Processing Systems (NIPS)},
  pages 1177--1184.

\bibitem[Ramachandran et~al., 2017]{ramachandran2017searching}
Ramachandran, P., Zoph, B., and Le, Q.~V. (2017).
\newblock Searching for activation functions.
\newblock {\em arXiv preprint arXiv:1710.05941}.

\bibitem[Romano and Wolf, 2005]{romano2005exact}
Romano, J.~P. and Wolf, M. (2005).
\newblock Exact and approximate stepdown methods for multiple hypothesis
  testing.
\newblock {\em Journal of the American Statistical Association},
  100(469):94--108.

\bibitem[Schrab et~al., 2022]{schrab2022ksd}
Schrab, A., Guedj, B., and Gretton, A. (2022).
\newblock {KSD} {A}ggregated goodness-of-fit test.
\newblock In {\em Advances in Neural Information Processing Systems 35: Annual
  Conference on Neural Information Processing Systems 2022, NeurIPS 2022}.

\bibitem[Schrab et~al., 2021]{schrab2021mmd}
Schrab, A., Kim, I., Albert, M., Laurent, B., Guedj, B., and Gretton, A.
  (2021).
\newblock {MMD} {A}ggregated two-sample test.
\newblock {\em arXiv preprint arXiv:2110.15073}.

\bibitem[Shao, 2010]{shao2010dependent}
Shao, X. (2010).
\newblock The dependent wild bootstrap.
\newblock {\em Journal of the American Statistical Association},
  105(489):218--235.

\bibitem[Song et~al., 2012]{song2012feature}
Song, L., Smola, A.~J., Gretton, A., Bedo, J., and Borgwardt, K.~M. (2012).
\newblock Feature selection via dependence maximization.
\newblock {\em Journal of Machine Learning Research}, 13:1393--1434.

\bibitem[Sriperumbudur et~al., 2011]{sriperumbudur2011universality}
Sriperumbudur, B.~K., Fukumizu, K., and Lanckriet, G.~R. (2011).
\newblock Universality, characteristic kernels and {RKHS} embedding of
  measures.
\newblock {\em Journal of Machine Learning Research}, 12(7).

\bibitem[Srivastava et~al., 2014]{srivastava2014dropout}
Srivastava, N., Hinton, G., Krizhevsky, A., Sutskever, I., and Salakhutdinov,
  R. (2014).
\newblock Dropout: a simple way to prevent neural networks from overfitting.
\newblock {\em Journal of Machine Learning Research}, 15(1):1929--1958.

\bibitem[Sutherland et~al., 2017]{sutherland2016generative}
Sutherland, D.~J., Tung, H.-Y., Strathmann, H., De, S., Ramdas, A., Smola, A.,
  and Gretton, A. (2017).
\newblock Generative models and model criticism via optimized maximum mean
  discrepancy.
\newblock In {\em International Conference on Learning Representations}.

\bibitem[Yamada et~al., 2018]{yamada2018post}
Yamada, M., Umezu, Y., Fukumizu, K., and Takeuchi, I. (2018).
\newblock Post selection inference with kernels.
\newblock In {\em International Conference on Artificial Intelligence and
  Statistics}, pages 152--160. PMLR.

\bibitem[Yamada et~al., 2019]{yamada2019post}
Yamada, M., Wu, D., Tsai, Y.~H., Ohta, H., Salakhutdinov, R., Takeuchi, I., and
  Fukumizu, K. (2019).
\newblock Post selection inference with incomplete maximum mean discrepancy
  estimator.
\newblock In {\em International Conference on Learning Representations}.

\bibitem[Zaremba et~al., 2013]{zaremba2013b}
Zaremba, W., Gretton, A., and Blaschko, M. (2013).
\newblock B-test: A non-parametric, low variance kernel two-sample test.
\newblock {\em Advances in neural information processing systems}, 26.

\bibitem[Zhang et~al., 2018]{zhang2018large}
Zhang, Q., Filippi, S., Gretton, A., and Sejdinovic, D. (2018).
\newblock Large-scale kernel methods for independence testing.
\newblock {\em Statistics and Computing}, 28(1):113--130.

\bibitem[Zhao and Meng, 2015]{zhao2015fastmmd}
Zhao, J. and Meng, D. (2015).
\newblock {FastMMD}: Ensemble of circular discrepancy for efficient two-sample
  test.
\newblock {\em Neural computation}, 27(6):1345--1372.

\end{thebibliography}

\clearpage
\appendix

\section*{Supplementary material for `Efficient Aggregated Kernel Tests using Incomplete $U$-statistics'}

\section{Assumptions}
\label{sec:assumptions}

\subsection{Assumptions for \Cref{theo:fixed} and \Cref{theo:indminimax}(i)}
\label{assump:fixed}
\begin{itemize}
    \item $\max\p{\|p\|_\infty,\|q\|_\infty}\leq M$ for some $M>0$
    \item $\alpha\in(0,e^{-1})$
    \item $\beta\in(0,1)$
    \item $B_1\geq 3\big(\!\log\!\big(8/\beta\big)+\alpha(1-\alpha)\big)\big/\alpha^2$
    \item $\sigma_{2,\lambda}^2\gtrsim 1$, where $\sigma_{2,\lambda}^2\coloneqq \E\!\left[h_\lambda(Z,Z')^2\right]$
\end{itemize}

\subsection{Assumptions for \Cref{theo:agg} and \Cref{theo:indminimax}(ii)}
\label{assump:agg}
\begin{itemize}
    \item $\max\p{\|p\|_\infty,\|q\|_\infty}\leq M$ for some $M>0$
    \item $\alpha\in(0,e^{-1})$
    \item $\beta\in(0,1)$
    \item $B_1\geq 12\Big(\underset{\lambda\in\Lambda}{\mathrm{max}}\, w_\lambda^{-2}\Big)\big(\log\p{8/\beta}+\alpha(1-\alpha)\big)\big/\alpha^2$
    \item $B_2\geq 8\log\!\big(2/\beta\big)\big/\alpha^2$
    \item $B_3 \geq \log_2\!\big(4\,\underset{\lambda\in\Lambda}{\mathrm{min}\, } w_\lambda^{-1} \big/\alpha\big)$
    \item $\sigma_{2,\lambda}^2\gtrsim 1$ for $\lambda\in\Lambda$, where $\sigma_{2,\lambda}^2\coloneqq \E\!\left[h_\lambda(Z,Z')^2\right]$
\end{itemize}

\section{Background details on MMD, HSIC, KSD and quantile estimation}
\label{sec:detailedbackground}

In this section, we present more background details than those presented in \Cref{sec:background} on the Maximum Mean Discrepancy, on the Hilbert Schmidt Independence Criterion, and on the Kernel Stein Discrepancy.

\textbf{Maximum Mean Discrepancy.} 
\citet{gretton2012kernel} introduce the {\em Maximum Mean Discrepancy} (MMD) which is a measure between probability densities $p$ and $q$ on $\R^d$.
It is defined as the integral probability metric \citep[IPM;][]{muller1997integral} over a reproducing kernel Hilbert space $\mathcal{H}_k$ \citep[RKHS;][]{azonszajn1950theory} with associated kernel $k$.
\citet[Lemma~4]{gretton2012kernel} show that the MMD is equal to the $\mathcal{H}_k$-norm of the difference between the mean embeddings 
$\mu_p(u) \coloneqq \mathbb{E}_{X\sim p}\left[k(X,u)\right]$
and $\mu_q(u) \coloneqq \mathbb{E}_{Y\sim q}\left[k(Y,u)\right]$ for $u\in\R^d$.
The square of the MMD is equal to 
\begin{align*}
    \mathrm{MMD}_k^2(p, q) 
    &\coloneqq \Big(\!\sup_{f\in\mathcal{H}_k \,:\, \norm{f}_{\mathcal{H}_k} \leq 1} \big|\mathbb{E}_p[f(X)] - \mathbb{E}_q [f(Y)]\big|\,\Big)^2 \\
    &= \norm{\mu_p-\mu_q}^2_{\mathcal{H}_k} \\
    &= \E_{p,p}\big[k\big(X,X'\big)\big] - 2\, \E_{p,q}\big[k\big(X,Y\big)\big] + \E_{q,q}\big[k\big(Y,Y'\big)\big]
\end{align*}
where $X$ and $X'$ (respectively $Y$ and $Y'$) are independent.
Using a characteristic kernel \citep{fukumizu2008kernel,sriperumbudur2011universality} guarantees that $\mathrm{MMD}_k^2(p, q) = 0$ if and only if $p=q$, a crucial property for using the MMD to construct a two-sample test.
With i.i.d.\ samples $\Xm \coloneqq (X_i)_{1\leq i\leq m}$ from $p$ and i.i.d.\ samples $\Yn= (Y_j)_{1\leq j\leq n}$ from $q$, \citet[Lemma~6]{gretton2012kernel} propose to use the unbiased quadratic-time MMD estimator $\widehat{\mathrm{MMD}}^2_k(\Xm,\Yn)$ defined as 
\begin{align*}
    &\frac{1}{m(m-1)} \sum_{(i,i')\in \textbf{i}_2^m} k(X_i,X_{i'})
    - \frac{2}{mn} \sum_{i=1}^m \sum_{j=1}^n k(X_i,Y_j)
    + \frac{1}{n(n-1)}  \sum_{(j,j')\in \textbf{i}_2^n} k(Y_j,Y_{j'}) \nonumber\\
    =\ &\frac{\one^\top\Kt_{\textrm{XX}}\one}{m(m-1)} 
    - 2 \frac{\one^\top \textbf{K}_{\textrm{XY}}\one}{mn}
    + \frac{\one^\top\Kt_{\textrm{YY}}\one}{n(n-1)} 
\end{align*}
where $\tilde{\textbf{K}}_{\textrm{XX}}$ and $\tilde{\textbf{K}}_{\textrm{YY}}$ are the kernel matrices $\textbf{K}_{\textrm{XX}}\coloneqq \big(k(X_i,X_j)\big)_{1\leq i,j\leq m}$ and $\textbf{K}_{\textrm{YY}}\coloneqq\big(k(Y_i,Y_j)\big)_{1\leq i,j\leq n}$ with diagonal entries set to 0, where $\textbf{K}_{\textrm{XY}}\coloneqq \big(k(X_i,Y_j)\big)_{1\leq i \leq m, 1\leq j\leq n}$, and where $\one$ is a one-dimensional vector with all entries equal to 1 of variable length determined by the context\footnote{We use this convention for the notation $\one$ in this whole section.}.
As noted by \citet{kim2020minimax}, this MMD estimator can be rewritten as a two-sample $U$-statistic (both of second order) \citep{hoeffding1992class} 
\begin{equation*}
    \widehat{\mathrm{MMD}}^2_k(\Xm,\Yn) = 
    \frac{1}{\abss{\textbf{i}_2^m}\abss{\textbf{i}_2^n}} 
    \sum_{(i,i')\in \textbf{i}_2^m}
    \sum_{(j,j')\in \textbf{i}_2^n}
    h_k^{\mathrm{MMD}}(X_i, X_{i'}; Y_j, Y_{j'})
\end{equation*}
where $\textbf{i}_a^b$ denotes the set of all $a$-tuples drawn without replacement from $\{1,\dots,b\}$ so that $\abss{\textbf{i}_a^b} = b \cdots (b-a+1)$, for example $\abss{\textbf{i}_2^m} = m(m-1)$, and where, for $x_1,x_2,y_1,y_2\in\R^d$, we let
\begin{equation*}
    h_k^{\mathrm{MMD}}(x_1, x_2; y_1, y_2) \coloneqq k(x_1,x_2)  - k(x_1,y_2) - k(x_2,y_1) + k(y_1,y_2).
\end{equation*}
This kernel can easily be symmetrized \citep{kim2020minimax} using a symmetrization trick \citep{dumbgen1998symmetrization}, this corresponds to working with
$$
    \bar h_k^{\mathrm{MMD}}(x_1, x_2; y_1, y_2) \coloneqq \frac{1}{2!2!} \sum_{(i_1,i_2)\in \textbf{i}_2^2} \sum_{(j_1,j_2)\in \textbf{i}_2^2} h_k^{\mathrm{MMD}}(x_1, x_2; y_1, y_2)
$$
and the MMD expression as a $U$-statistic still holds when replacing $h_k^{\mathrm{MMD}}$ with its symmetrized variant $\bar h_k^{\mathrm{MMD}}$.

\textbf{Hilbert Schmidt Independence Criterion.}
For a joint probability density $p_{xy}$ on $\R^{d_x}\times\R^{d_y}$ with marginals $p_x$ on $\R^{d_x}$ and $p_y$ on $\R^{d_y}$,
\citet{gretton2005kernel} introduce the {\em Hilbert Schmidt Independence Criterion} (HSIC) which is defined as
\begin{align*}
    \mathrm{HSIC}_{k,\ell}(p_{xy}) 
    &\coloneqq \mathrm{MMD}_\kappa^2(p_{xy}, p_xp_y) \\
    &= \E_{p_{xy},p_{xy}}\Big[k(X,X')\ell(Y,Y')\Big] - 2\, \E_{p_{xy}}\Big[\E_{p_{x}}[k(X,X')]\E_{p_{y}}[\ell(Y,Y')]\Big] \\
    &\hspace{4.535cm}+ \E_{p_x,p_x}\Big[k(X,X')\Big] \E_{p_y,p_y}\Big[k(Y,Y')\Big].
\end{align*}
with kernels $k$ on $\R^{d_x}$ and $\ell$ on $\R^{d_y}$ giving the product kernel $\kappa\big((x,y), (x',y')\big) \coloneqq k(x,x')\ell(y,y')$ on $\R^{d_x}\times\R^{d_y}$.
With i.i.d.\ pairs of samples $\Zn \coloneqq \big(Z_i\big)_{1\leq i\leq N} = \big((X_i,Y_i)\big)_{1\leq i\leq N}$ drawn from $p_{xy}$, a natural unbiased HSIC estimator \citep{gretton2008kernel,song2012feature} is then
\begin{align*}
    \widehat{\mathrm{HSIC}}_{k,\ell}(\Zn) 
    &\coloneqq
    \frac{1}{\abss{\textbf{i}_2^N}} 
    \sum_{(i,j)\in \textbf{i}_2^N}
    k(X_i, X_j) \ell(Y_i, Y_j)
    - 
    \frac{2}{\abss{\textbf{i}_3^N}} 
    \sum_{(i,j,r)\in \textbf{i}_3^N}
    k(X_i, X_j) \ell(Y_i, Y_r) \nonumber  \\
    &\hspace{5.089cm}+
    \frac{1}{\big|\textbf{i}_4^N\big|} 
    \sum_{(i,j,r,s)\in \textbf{i}_4^N}
    k(X_i, X_j) \ell(Y_r, Y_s) \nonumber \\
    &= 
    \frac{1}{\big|\textbf{i}_4^N\big|} 
    \sum_{(i,j,r,s)\in \textbf{i}_4^N}
    h_{k,\ell}^{\mathrm{HSIC}}(Z_i, Z_j, Z_r, Z_s)
\end{align*}
which is a fourth-order one-sample $U$-statistic.
For $z_a=(x_a,y_a)\in\R^{d_x}\times\R^{d_y}$, $a=1,\dots,4$, we let
\begin{equation*}
    h_{k,\ell}^{\mathrm{HSIC}}(z_1, z_2, z_3, z_4)
    \coloneqq
    \frac{1}{4}
    h_k^{\mathrm{MMD}}(x_1, x_2; x_3, x_4)
    h_\ell^{\mathrm{MMD}}(y_1, y_2; y_3, y_4)
    .
\end{equation*}
We stress the fact that this HSIC estimator can actually be computed in quadratic time as shown by \citet[Equation 5]{song2012feature} who provide the following closed-form expression
\begin{equation*}
    \widehat{\mathrm{HSIC}}_{k,\ell}(\Zn) 
    = \frac{1}{N(N-3)}
    \left(
    \textrm{tr}(\Kt\Lt)
    +
    \frac
    {\one^\top\Kt\one\one^\top\Lt\one}
    {(N-1)(N-2)}
    -
    \frac{2}{N-2}
    \one^\top\Kt\Lt\one
    \right)
\end{equation*}
where $\Kt$ and $\Lt$ are the kernel matrices $\textbf{K}\coloneqq \big(k(X_i,X_j)\big)_{1\leq i,j\leq N}$ and $\textbf{L}\coloneqq\big(\ell(Y_i,Y_j)\big)_{1\leq i,j\leq N}$ with diagonal entries set to 0.
Again, this kernel can be symmetrized \citep{song2012feature,kim2020minimax} using a symmetrization trick \citep{dumbgen1998symmetrization}, and the HSIC expression as a $U$-statistic still holds when replacing $h_k^{\mathrm{HSIC}}$ with its symmetrized variant
$$
    \bar h_k^{\mathrm{HSIC}}(z_1, z_2, z_3, z_4) \coloneqq \frac{1}{4!} \sum_{(i_1,i_2,i_3,i_4)\in \textbf{i}_4^4} h_k^{\mathrm{HSIC}}(z_{i_1}, z_{i_2}, z_{i_3}, z_{i_4}).
$$

\textbf{Kernel Stein Discrepancy.}
For probability densities $p$ and $q$ on $\R^d$, \citet{chwialkowski2016kernel} and \citet{liu2016kernelized} introduce the {\em Kernel Stein Discrepancy} (KSD) defined as 
\begin{align*}
    \mathrm{KSD}_{p,k}^2(q) 
    &\coloneqq \mathrm{MMD}^2_{h_{k,p}^{\mathrm{KSD}}}(q,p) \\
    &= \E_{q,q}\big[h_{k,p}^{\mathrm{KSD}}\big(Z,Z'\big)\big] 
    - 2\, \E_{q,p}\big[h_{k,p}^{\mathrm{KSD}}\big(Z,X\big)\big] 
    + \E_{p,p}\big[h_{k,p}^{\mathrm{KSD}}\big(X,X'\big)\big] \\
    &= \E_{q,q}\big[h_{k,p}^{\mathrm{KSD}}\big(Z,Z'\big)\big]
\end{align*}
where the {\em Stein kernel} $h_{p,k}\colon\R^d\times\R^d\to\R$ is defined as
\begin{equation*}
\begin{aligned}
    h_{k,p}^{\mathrm{KSD}}(x,y) \coloneqq\ 
    &\p{\nabla\log p(x)^\top \nabla\log p(y)} k(x,y)
    + \nabla\log p(y)^\top \nabla_x k(x,y) \\
    &+ \nabla\log p(x)^\top \nabla_y k(x,y)
    + \sum_{i=1}^d \frac{\partial}{\partial x_i \partial y_i}\, k(x,y).
\end{aligned}
\end{equation*}
The Stein kernel satisfies the Stein identity $\E_{p}[h_{k,p}^{\mathrm{KSD}}(X,\cdot)] = 0$.
The KSD is particularly useful for the goodness-of-fit setting with a model density $p$ and i.i.d.\ samples $\Zn \coloneqq (Z_i)_{1\leq i\leq N}$ drawn from a density $q$ because it admits an estimator which does not require samples from the model $p$.
The quadratic-time KSD estimator can be computed as the second-order one-sample $U$-statistic
\begin{equation*}
    \widehat{\mathrm{KSD}}_{p,k}^2(\Zn)
    \coloneqq 
    \frac{1}{\abss{\textbf{i}_2^N}} 
    \sum_{(i,j)\in \textbf{i}_2^N}
    h_{k,p}^{\mathrm{KSD}}(Z_i,Z_j)
    = \frac{\one^\top \tilde{\textbf{H}} \one}{N(N-1)}
\end{equation*}
where $\tilde{\textbf{H}}$ is the kernel matrix $\textbf{H}\coloneqq \big(h_{k,p}^{\mathrm{KSD}}(Z_i,Z_j)\big)_{1\leq i,j\leq N}$ with diagonal entries set to 0.
The Stein kernel $h_k^{\mathrm{KSD}}$ is already symmetric, we can write $\bar h_k^{\mathrm{KSD}}(x,y) \coloneqq h_k^{\mathrm{KSD}}(x,y)$ for all $x,y\in\R^d$ for consistency of notation.
As presented in \Cref{sec:background}, \citet[Theorem 2.2]{chwialkowski2016kernel} show the consistency of the KSD goodness-of-fit provided that the kernel $k$ is $C_0$-universal \citep[Definition 4.1]{carmeli2010vector} and that 
\begin{equation*}
    \mathbb{E}_q \!\Big[h_{k,p}^{\mathrm{KSD}}(z,z) \Big]<\infty    
    \qquad \qquad \text{and} \qquad \qquad 
    \mathbb{E}_q \!\Bigg[\left\|\nabla \log\p{\frac{p(z)}{q(z)}}\right\|_2^2\Bigg]<\infty.
\end{equation*}
as introduced in \Cref{eq:ksdassumptions}.

\textbf{Quantile estimation.}
There exist many approaches to estimate the quantiles of the test statistics under the null hypothesis in the three frameworks:
using the quantile of a known distribution-free asymptotic null distribution \citep{gretton2008kernel,gretton2012kernel},
sampling from an asymptotic null distribution with eigenspectrum approximation \citep{gretton2009fast},
using permutations \citep{gretton2008kernel,albert2019adaptive,kim2020minimax,schrab2021mmd},
using a wild bootstrap \citep{fromont2012kernels,chwialkowski2014wild,chwialkowski2016kernel,schrab2021mmd,schrab2022ksd},
using a parametric bootstrap \citep{key2021composite,schrab2022ksd},
using other bootstrap methods \citep{liu2016kernelized},
to name but a few.
Permutation-based tests have been shown to correctly control the non-asymptotic level for the two-sample \citep{schrab2021mmd,kim2020minimax} and independence \citep{albert2019adaptive,kim2020minimax} problems.
For the two-sample test, using a wild bootstrap also guarantees well-calibrated non-asymptotic level \citep{fromont2012kernels,schrab2021mmd}.
For the goodness-of-fit setting, while a wild bootstrap guarantees only control of the asymptotic level \citep{chwialkowski2016kernel}, using a parametric bootstrap 
results in a well-calibrated non-asymptotic level \citep{schrab2022ksd}.
In this work, we focus on the wild bootstrap approach, though we point out that our results also hold using a parametric bootstrap for the goodness-of-fit setting as done by \citet{schrab2022ksd}.

\section{Detailed experimental protocol}
\label{sec:detailedexperiments}

In this section, we present details on our experiments and on the tests considered.

\textbf{Implementation and computational resources.}
All experiments have been run on an AMD Ryzen Threadripper 3960X 24 Cores 128Gb RAM CPU at 3.8GHz, except the LSD test \citep{grathwohl2020learning} for which a neural network has been trained using an NVIDIA RTX A5000 24Gb Graphics Card.
The overall runtime of all the experiments is of the order of a couple of hours (significant speedup can be obtained by using parallel computing).
We use the implementations of the respective authors (all under the MIT license)
for the \href{https://github.com/wittawatj/interpretable-test}{ME}, \href{https://github.com/wittawatj/interpretable-test}{SCF}, \href{https://github.com/wittawatj/fsic-test}{FSIC} and \href{https://github.com/wittawatj/kernel-gof}{FSSD} tests of \citet{jitkrittum2016interpretable,jitkrittum2017adaptive,jitkrittum2017linear}, 
for the \href{https://github.com/wgrathwohl/LSD}{LSD} test of \citet{grathwohl2020learning},
for the \href{https://bitbucket.org/jhhuggins/random-feature-stein-discrepancies}{L1 IMQ and Cauchy RFF} tests of \citet{huggins2018random},
and for the \href{https://github.com/jmkuebler/tests-wo-splitting}{OST PSI} test of \citet{kubler2020learning}.
The implementation of our computationally efficient aggregated tests, as well as the code for reproducibility of the experiments, are available \href{https://github.com/antoninschrab/agginc-paper}{here} under the MIT license.

\textbf{Kernels.}
For the two-sample and independence experiments, we use the Gaussian kernel\footnote{In practice, we do not need to normalize the kernels to integrate to 1 since our tests are invariant to multiplying the kernel by a scalar.} with equal bandwidths $\lambda_1=\dots=\lambda_d = \tilde\lambda$, which is defined as
$$
    k_\lambda(x,y) \coloneqq  \exp\p{-\sum_{i=1}^d\frac{(x_i-y_i)^2}{\lambda_i^2}}
    = \exp\p{-\frac{\norm{x-y}_2^2}{\tilde \lambda^2}},
$$
and similarly for the kernel $\ell_\mu$.
As shown by \citet{gorham2017measuring}, a more appropriate kernel for goodness-of-fit testing is the IMQ (inverse multiquadric) kernel
\begin{equation}
    \label{eq:imqkernel}
    k_\lambda(x,y) \coloneqq 
    \p{1+\sum_{i=1}^d\frac{(x_i-y_i)^2}{\lambda_i^2}}^{-\beta_k}
    = \p{1+\frac{\norm{x-y}_2^2}{\tilde \lambda^2}}^{-\beta_k}
    \propto \p{\tilde \lambda^2+\norm{x-y}_2^2}^{-\beta_k}
\end{equation}
for some $\beta_k\in(0,1)$.
In our goodness-of-fit experiments, we use the IMQ kernel with fixed parameter $\beta_k = 0.5$.

\textbf{Two-sample and independence experiments.}
In our experiments, we consider perturbed uniform densities, those can be shown to lie in Sobolev balls and are used to derive the minimax rates over Sobolev balls for the two-sample and independence problems \citep{li2019optimality,albert2019adaptive}.
For the two-sample problem, we consider testing samples drawn from a uniform density against samples drawn from a perturbed uniform density, as considered by \citet[see Equation 17 for formal definition and Figure 2 for illustrations]{schrab2021mmd}.
We scale the perturbations so that the perturbed density takes value in the whole interval $[0,2]$, we then consider some inverse scaling parameter $S\geq 1$ such that it takes value in the interval $[1-1/S, 1+1/S]$. Intuitively, as $S$ increases, the perturbation is shrunk.
In \Cref{fig:experiments} (a, d), we consider 2 perturbations with inverse scaling parameter $S=2$ in dimension $d=1$ and vary the sample size $N\in\{200, 400, 600, 800, 1000\}$.
In \Cref{fig:experiments} (b), we vary the dimension $d\in\{1,2,3,4\}$ for 1 perturbation with $S=1$ and $N=1000$.
In \Cref{fig:experiments} (c), we use 1 perturbation with $d=1$ and $N=1000$, we vary the inverse scaling parameter $S\in\{1,2,3,4,5\}$.
For the independence problem, we draw samples from the joint perturbed uniform density in dimension $d_x+d_y$, the marginals are simply uniform densities in dimensions $d_x$ and $d_y$, respectively.
We fix $d_x=1$ and vary $d_y$ exactly as in the two-sample setting. 
The parameters for the independence experiments in \Cref{fig:experiments} (e--h) are the same as those of the two-sample experiments in \Cref{fig:experiments} (a--d) detailed above (with the only difference that for \Cref{fig:experiments} (f) we consider $d_y\in\{1,2,3\}$).

\textbf{Goodness-of-fit experiments.}
In \Cref{fig:experiments} (i--l), we use a Gaussian-Bernoulli Restricted Boltzmann Machine (GBRBM) with the same setting considered by \citet{liu2016kernelized}, \citet{grathwohl2020learning} and \citet{schrab2022ksd}.
This is a hidden variable model with a continuous observable variable in $\R^{d_x}$ and a hidden binary variable in $\{-1,1\}^{d_h}$, the joint density is intractable but the score function admits a closed form.
The GBRBM has parameters $b\in\R^{d_x}$ and $c\in\R^{d_h}$, which are drawn from Gaussian standard distributions, and a matrix parameter $B\in\R^{d_x \times d_h}$.
For the model $p$, the elements of $B$ are sampled uniformly from $\{-1,1\}$ (i.i.d.\ Rademacher variables).
The samples come from a GBRBM $q$ with the same parameters as the model $p$ but where some Gaussian noise $\mathcal{N}(0,\sigma)$ is injected into the elements of $B$.
In \Cref{fig:experiments} (i, l), we consider dimensions $d_x=50$ and $d_h=40$ with noise standard deviation $\sigma=0.02$ and we vary the sample size $N\in\{200, 400, 600, 800, 1000\}$.
In \Cref{fig:experiments} (j), we fix $d_x=100$, $N=1000$, $\sigma=0.03$ and we vary the hidden dimension $d_h\in\{20,40,60,80\}$.
For fixed observed dimension $d_x$, as the hidden dimension $d_h$ increases the size of $B\in\R^{d_x \times d_h}$ becomes larger, so there is more evidence of the noise being injected, which makes the problem easier. 
Hence, the test power increases as $d_h$ increases for fixed $d_x$.
In \Cref{fig:experiments} (k), we consider dimensions $d_x=50$ and $d_h=40$ with sample size $N=1000$, we vary the noise standard deviations $\sigma\in\{0, 0.01, 0.02, 0.03, 0.04\}$.

\textbf{AggInc tests.}
As in \citet{schrab2021mmd}, for MMDAggInc, we use a collection of $B=10$ bandwidths defined as 
    $$
        \left\{
            \left(4\lambda_{\textrm{max}}/\lambda_{\textrm{min}}\right)^{(i-1) / (B - 1)} : i = 1, \dots, B
        \right\}
    $$
which is a discretisation of the interval $\left[\lambda_{\textrm{min}}/2,2\lambda_{\textrm{max}}\right]$ where $\lambda_{\textrm{min}}$ and $\lambda_{\textrm{max}}$ are the minimal and maximal inter-sample distances, respectively. 
If the minimal distance is smaller than $10^{-1}$, we consider the 5\% smallest inter-sample positive distance instead, if it is still smaller than $10^{-1}$ we set $\lambda_{\textrm{min}}=10^{-1}$.

Similarly to \citet{schrab2022ksd}, for KSDAggInc, we use a collection of $B=10$ bandwidths defined as 
    $$
        \left\{
            d^{-1}\lambda_{\textrm{max}}^{(i-1) / (B - 1)} : i = 1, \dots, B
        \right\}
    $$
where $d$ is the dimension of the samples, and where $\lambda_{\textrm{max}}=\max\big\{\norm{z_i-z_j}_2:(i,j)\in \mathbf{i}_2^N\big\}$ is the maximal inter-sample distance (which is thresholded at 2 if it is smaller than this value).
This collection discretises the interval $\left[d^{-1}, d^{-1}\lambda_{\textrm{max}}\right]$.

For HSICAggInc, we work with the collection of 25 pairs of bandwidths
$$
    \Lambda \coloneqq \Big\{\p{2^i\lambda_{\textrm{med}}\one_{d_x}, 2^j\mu_{\textrm{med}}\one_{d_y}} : i,j \in\{-2,-1,0,1,2\}\Big\}
$$
for the kernels $k_\lambda$ and $\ell_\mu$ defined in \Cref{kernels},
where
$$
    \lambda_{\textrm{med}} \coloneqq \textrm{median}\Big\{\norm{x_i-x_j}_2: (i,j)\in \mathbf{i}_2^N\Big\}
    \quad \textrm{ and } \quad
    \mu_{\textrm{med}} \coloneqq \textrm{median}\Big\{\norm{y_i-y_j}_2: (i,j)\in \mathbf{i}_2^N\Big\}.
$$
We also consider collections of this form for MMDAggInc and KSDAggInc in \Cref{sec:experiments_old} but this results in a small loss of power.

In practice, depending on our computational budget, we can also consider multiple kernels, each with various bandwidths, as considered in \citet{schrab2021mmd}.
All aggregated tests are run with uniform weights defined as $w_\lambda \coloneqq 1/ \abs{\Lambda}$ for all $\lambda\in\Lambda$.
The design choice consists of $R$ sub-diagonals of the kernel matrix for $R\in\{1,100,200\}$, it is formally defined in \Cref{sec:experiments}.
We also consider the quadratic-time case where the full design is considered (\emph{i.e.} case $R=N-1$), we refer to these tests using complete $U$-statistics as AggCom for consistency.
We note that MMDAggCom, HSICAggCom and KSDAggCom simply correspond to the quadratic-time MMDAgg, HSICAgg and KSDAgg tests proposed by \citet{schrab2021mmd}, \citet{albert2019adaptive} and \citet{schrab2022ksd}, respectively, with the only difference being their implementation: Agg tests run slightly faster than AggCom tests since they can exploit the fact that the whole kernel matrix needs to be computed.
We use $B_1= 500$ and $B_2= 500$ wild bootstrapped statistics to estimate the quantiles and the probability under the null for the correction in \Cref{eq:correction}, respectively.
In practice, we recommend using either $B_1=B_2=500$ for having fast tests, or $B_1=B_2=2000$ for obtaining slightly higher power (with longer runtimes).
For that correction term, we use $B_3=50$ steps of bisection method to approximate the supremum.

\textbf{ME, SCF, FSIC and FSSD tests.} 
\citet{jitkrittum2016interpretable} use the two-sample tests ME and SCF proposed by \citet{chwialkowski2015fast} with features which are chosen to maximise a lower bound on the test power.
The ME test is based on analytic Mean Embeddings while the SCF test uses the difference in Smooth Characteristic Functions.
For the independence problem, \citet{jitkrittum2017adaptive} construct a FSIC test which uses their proposed normalised Finite Set Independence Criterion.
\citet{jitkrittum2017linear} propose a goodness-of-fit test based on the Finite Set Stein Discrepancy (FSSD).
All those tests utilise test statistics which evaluate the witness function of either the MMD, HSIC, or KSD, at some test locations (\emph{i.e.} features) chosen on held-out data to maximise test power.
For the two-sample SCF test, the test locations are in the frequency domain rather than in the spatial domain.
All tests are used with 10 test locations which are chosen on half of the data, as done in the experiments of \citet{jitkrittum2016interpretable,jitkrittum2017adaptive,jitkrittum2017linear}. 
The ME and SCF tests use the quantiles of their known chi-square asymptotic null distributions.
The FSIC test uses 500 permutations to simulate the null hypothesis and compute the test threshold to ensure a well-calibrated non-asymptotic level.
The FSSD test simulates 2000 samples from the asymptotic null distribution (weighted sum of chi-squares) with the eigenvalues being computed from the covariance matrix with respect to the observed samples.
For the two-sample and independence tests, the bandwidths of the Gaussian kernels are selected during the optimization procedure.
For the goodness-of-fit test, the bandwidth of the IMQ (inverse multiquadric) kernel is set to some fixed value as done by \citet{jitkrittum2017linear}, following the recommendation of \citet{gorham2017measuring}.

\textbf{LSD test.}
The Kernelised Stein Discrepancy (KSD) is a Stein Discrepancy \citep{gorham2017measuring} where the class of functions is taken to be the unit ball of a reproducing kernel Hilbert space (RKHS).
\citet{grathwohl2020learning} propose to instead consider some more expressive class of functions consisting of neural networks, resulting in the Learned Stein Discrepancy (LSD).
For goodness-of-fit testing, they propose to split the data into training (80\%), validation (10\%) and testing (10\%) sets.
They construct a test statistic which is asymptotically normal under both $\mathcal H_0$ and $\mathcal H_1$.
Using the training set, they train the parametrised neural network to maximise the test power by optimizing a proxy for it which is derived following the reasonings of \citet{gretton2012optimal}, \citet{sutherland2016generative} and \citet{jitkrittum2017linear}.
They perform model selection on the validation set. 
Finally, they run the test on the testing set using the quantile of the asymptotic normal distribution under the null.
As in the experiments of \citet{grathwohl2020learning}, a 2-layer MLP with 300 units per layer and with Swish nonlinearity \citep{ramachandran2017searching} is used.
Their model is trained using the Adam optimizer \citep{kingma2014adam} for 1000 iterations, with dropout \citep{srivastava2014dropout}, with weight decay of strength $0.0005$, with learning rate $10^{-3}$, and with $L^2$ regularising strength $0.5$.

\textbf{Cauchy RFF and L1 IMQ tests.} 
\citet{huggins2018random} introduce random feature Stein discrepancies (R$\Phi$SDs) which are computable in linear time. 
The FSSD of \citet{jitkrittum2017linear} corresponds to a specific R$\Phi$SD.
Another special case of their general R$\Phi$SDs is the random Fourier feature \citep[RFF;][]{rahimi2007random} approximation of KSD. 
They consider in their experiments both Gaussian and Cauchy RFF tests, they observe that Cauchy RFF significantly outperforms its Gaussian counterpart \citep[Figure 4]{huggins2018random}.
Using the inverse multiquadric kernel (IMQ; \Cref{eq:imqkernel}), for which \citet{gorham2017measuring} showed that KSD dominates weak convergence when $\beta_k\in(0,1)$, \citet[Example 3.4]{huggins2018random} derive a $L^r$ IMQ R$\Phi$SD, with some simple setting when $r=1$.
They show in their experiments that L1 IMQ has superior performance compared to all other tests considered for experiments comparing Gaussian and Laplace distributions, as well as Gaussian and multivariate $t$ distributions.
We use the parameters recommended by the authors when running Cauchy RFF and L1 IMQ, 
except for the number of samples drawn from the unnormalized density to estimate the covariance matrix to simulate the null hypothesis.
As explained in \Cref{sec:ndraws}, we tune that number in order for their tests to be more computationally efficient while retaining their high test power.

\textbf{OST PSI test.} 
\citet{kubler2020learning} construct an MMD adaptive two-sample test which exploits the post-selection inference framework \citep[PSI;][]{fithian2014optimal,lee2016exact} (with uncountable candidate sets) to use the same data to both perform kernel selection and run the test while still guaranteeing control of the probability of type I error.
Their one-sided test (OST) runs in linear time and does not rely on data splitting. 
For kernel selection, they use a proxy for asymptotic power as a criterion. 
We use their implementation with the same collection of bandwidths as for our MMDAggInc test as specified above.

\section{Additional experiments}
\label{sec:additional_experiments}

In this section, we present additional experiments. 
We consider more challenging experiments on the high-dimensional MNIST dataset. 
We report results using different collections of bandwidths.
We empirically show that all the tests considered have well-calibrated levels.
We present experiments highlighting the strengths of the aggregation procedure.
Finally, we discuss the choice of parameters for the L1 IMQ and Cauchy RFF tests.

\subsection{MNIST Experiments}

\begin{figure}[t]
  \centering
  \includegraphics[width=\textwidth]{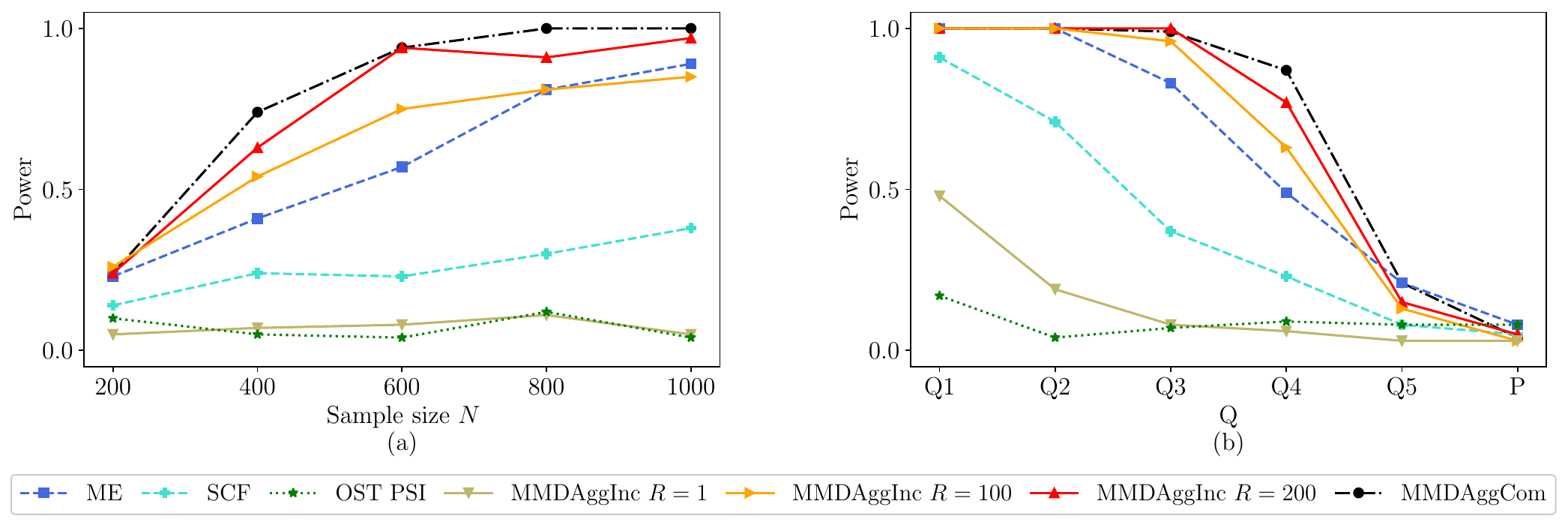}
  \includegraphics[width=\textwidth]{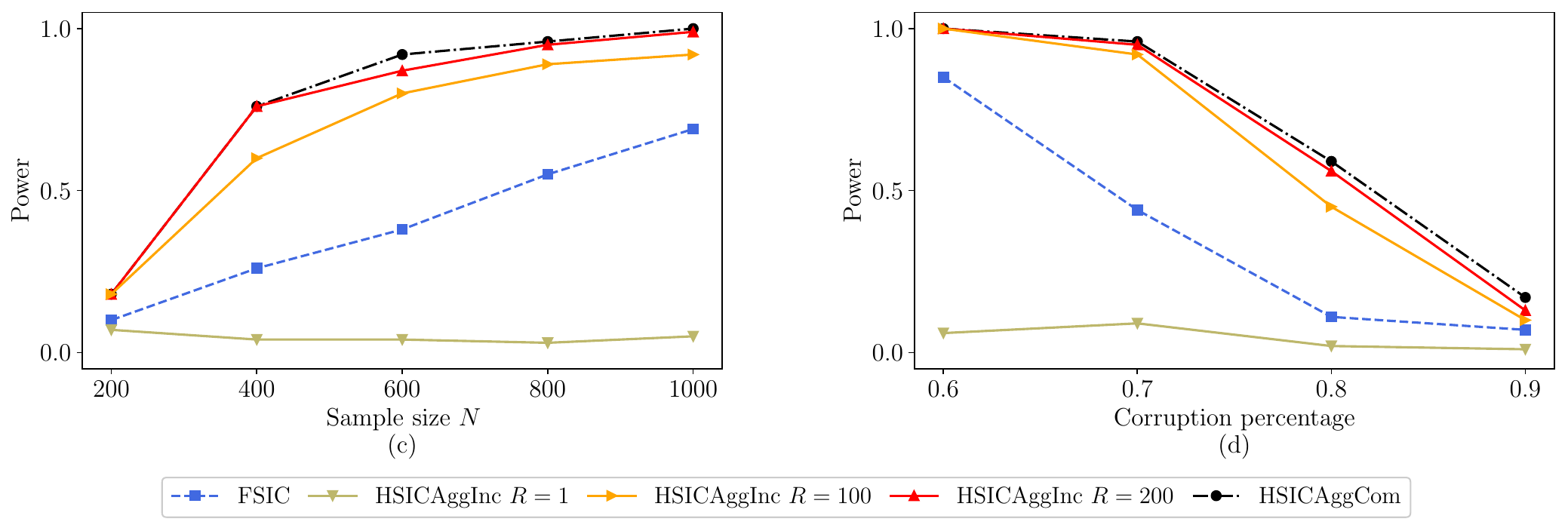}
  \includegraphics[width=0.7\textwidth]{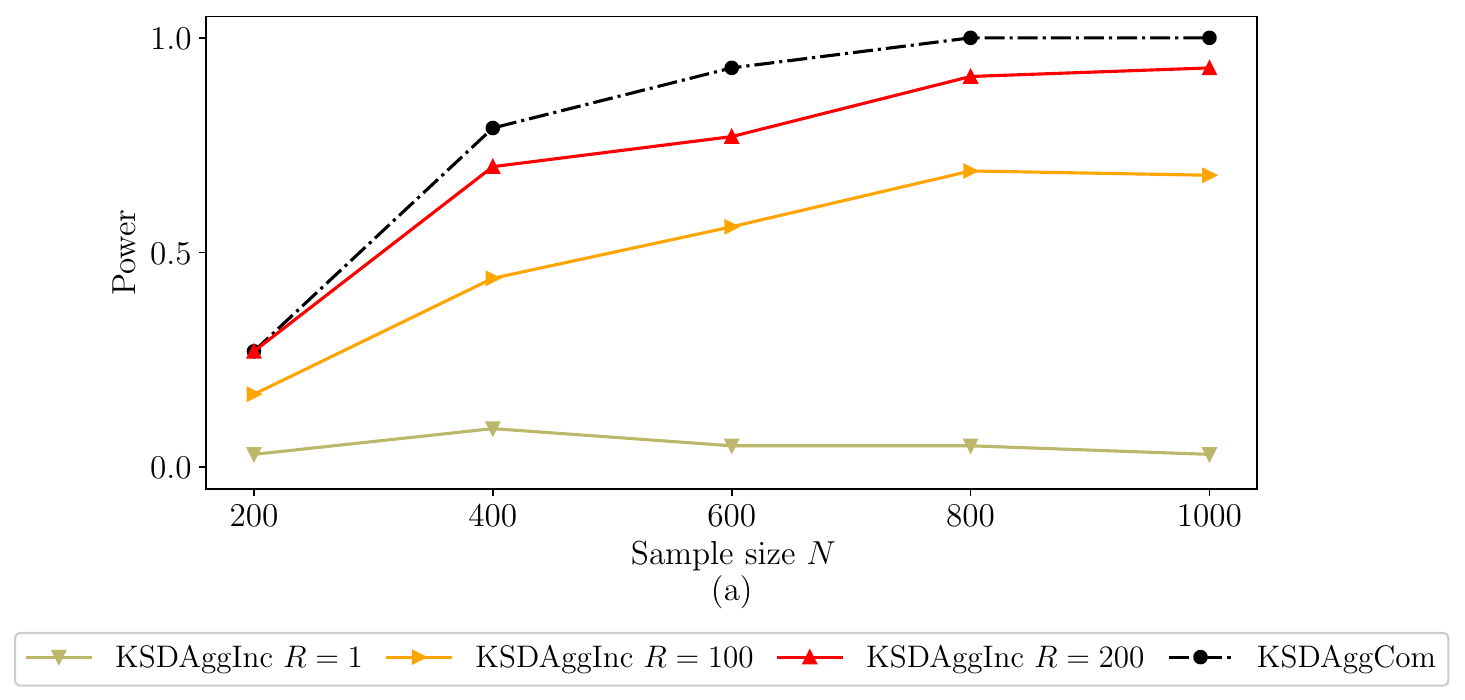}
  \caption{%
  Experiments using the MNIST dataset for the {\em (a--b)} two-sample, {\em (c--d)} independence, and {\em (e)} goodness-of-fit problems. 
  The power results are averaged over 100 repetitions.
  \label{fig:experiments_mnist}
  }
\end{figure}

In \Cref{fig:experiments_mnist}, we run experiments on the real-world MNIST dataset \citep{lecun2010mnist} consisting of images of digits in dimension 784. 

For the two-sample problem, the distribution $P$ consists of images of all digits and the other distribution is $Q_i$ where $Q_1$ consists of images of only the five odd digits, 
$Q_2$ is $Q_1$ with $0$, 
$Q_3$ is $Q_2$ with $2$, 
$Q_4$ is $Q_3$ with $4$, 
$Q_5$ is $Q_4$ with $6$ (\emph{i.e.} $Q_5$ consists of images of all digits expect 8).
This setting has previously been considered by \cite{schrab2021mmd}.

For the independence problem, we pair each image of a digit with the value of the digit. 
To make the problem more challenging, we corrupt some percentage of the data by pairing images with values of random digits.

For the goodness-of-fit problem, the samples are drawn from the true MNIST dataset and the model is a Normalizing Flow \citep[generative model which admits a density;][]{dinh2016density,kingma2018glow} trained on the MNIST dataset. 
Since we have access only to pre-computed values of the score function evaluated at some MNIST samples but do not have access to the score function itself, we found that computing FSSD, L1 IMQ or Cauchy RFF to be very challenging; for this reason the results for those tests are not reported.

Overall, we observe the same trends in \Cref{fig:experiments_mnist} for this high-dimensional real-world setting as we did in \Cref{fig:experiments} in the lower-dimensional setting in which the Sobolev smoothness assumption is satisfied for MMDAggInc and HSICAggInc.
Indeed, the AggInc $R=200$ tests clearly outperform the tests we compare against and even match the power of AggCom in several experiments.
ME and SCF obtain significantly lower power than MMDAggInc $R=100$ in various settings in \Cref{fig:experiments_mnist} (a, b).
We observe that HSICAggInc $R=100$ significantly outperforms FSIC in both independence experiments in \Cref{fig:experiments_mnist} (c, d).
For the goodness-of-fit setting, the tests manage to detect that the true MNIST samples are not drawn from the density of the trained Normalizing Flow.
There is a significant power difference between each of the four tests: KSDAggInc $R=1,100,200$ and KSDAggCom.

\FloatBarrier

\subsection{Different collections for MMDAggInc and KSDAggInc}
\label{sec:experiments_old}

\begin{figure}
  \centering
  \includegraphics[width=\textwidth]{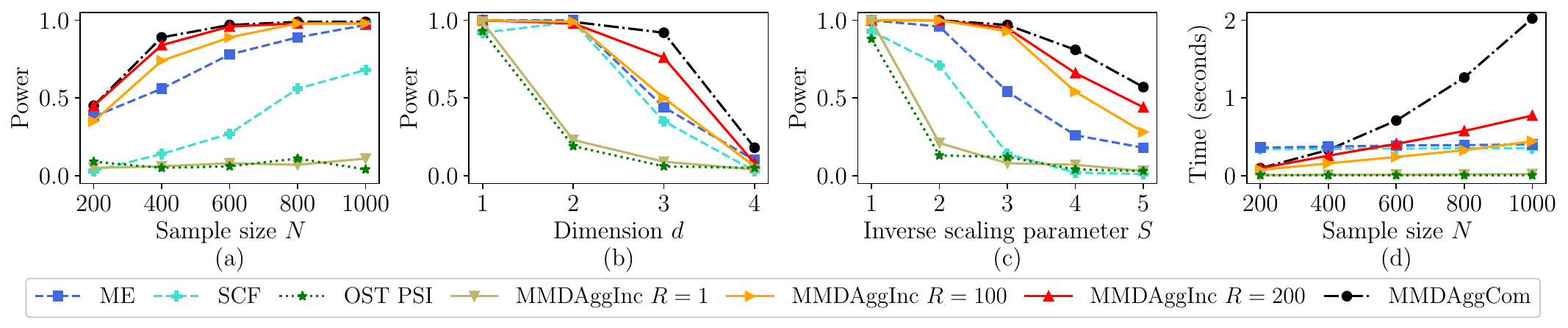}
  \includegraphics[width=\textwidth]{figures/figure_hsic-eps-converted-to.pdf}
  \includegraphics[width=\textwidth]{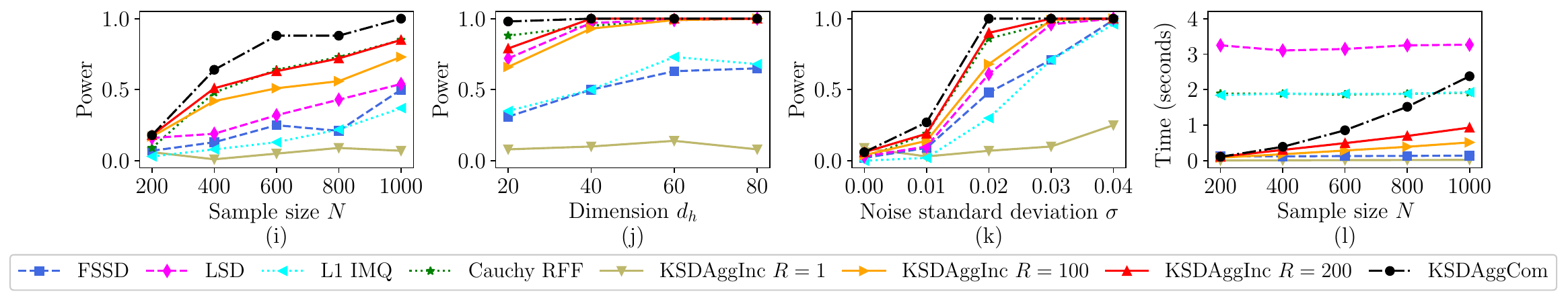}
  \caption{%
  Two-sample {\em (a--d)} and independence {\em (e--h)} experiments using perturbed uniform densities.
  Goodness-of-fit {\em (i--l)} experiment using a Gaussian-Bernoulli Restricted Boltzmann Machine.
  The power results are averaged over 100 repetitions and the runtimes over 20 repetitions. 
  \label{fig:experiments_old}
  }
\end{figure}

In \Cref{fig:experiments_old}, we reproduce the experiments presented in \Cref{fig:experiments} using, for MMDAggInc and KSDAggInc, the collection of 21 bandwidths
$$
    \Lambda \coloneqq \Big\{2^i\lambda_{\textrm{med}}\one_d: i \in\{-10,\dots,10\} \Big\}
    \quad \textrm{ where } \quad
    \lambda_{\textrm{med}} \coloneqq \textrm{median}\Big\{\norm{z_i-z_j}_2: (i,j)\in \mathbf{i}_2^N\Big\},
$$
where $\one_d$ is a $d$-dimensional vector with all entries equal to 1. 
We observe that using this collection leads to slightly lower power for MMDAggInc and KSDAggInc than in \Cref{fig:experiments} with different collections. 
In \Cref{fig:experiments_old}, KSDAggInc $R=200$ obtains exactly the same power as Cauchy RFF.
The results for HSICAggInc in \Cref{fig:experiments_old} are the same as those of \Cref{fig:experiments}, we simply report them for consistency.

\FloatBarrier

\subsection{Well-calibrated levels}

All tests are run with level $\alpha = 0.05$, it is verified in \Cref{tab:mmdn,tab:mmdd,tab:hsicn,tab:hsicd,tab:ksdn,tab:ksdd} that all tests have well-calibrated levels for the three testing frameworks, when varying either the sample size or the dimension.
The levels plotted are averages obtained across 200 repetitions, this explains the small fluctuations observed from the desired test level $\alpha = 0.05$.
The settings of those six experiments correspond to the settings of the experiments presented in \Cref{fig:experiments} (a, b, e, f, i, j) detailed above, with the difference that we are working under the null hypothesis (\emph{i.e.}\ perturbed uniform densities are replaced with uniform densities, and the noise standard deviation for the Gaussian-Bernoulli Restricted Boltzmann Machine is set to $\sigma=0$). 

\begin{table}[h!]
\caption{Two-sample level experiment using uniform densities varying the sample size.}
\label{tab:mmdn}
\vspace{-0.3cm}
\begin{center}
\begin{small}
\begin{tabular}{cccccccc}
\toprule
\begin{tabular}{@{}c@{}}Sample \\ size\end{tabular} 
& ME 
& SCF 
& \begin{tabular}{@{}c@{}}OST \\ PSI\end{tabular}  
& \begin{tabular}{@{}c@{}}MMDAggInc \\ $R=1$\end{tabular}  
& \begin{tabular}{@{}c@{}}MMDAggInc \\ $R=100$\end{tabular}
& \begin{tabular}{@{}c@{}}MMDAggInc \\ $R=200$\end{tabular}
& MMDAggCom
\\
\midrule
200 & 0.055 & 0.005 & 0.045 & 0.04 & 0.05 & 0.055 & 0.055 \\
400 & 0.08 & 0.01 & 0.04 & 0.035 & 0.06 & 0.03 & 0.03 \\
600 & 0.08 & 0.005 & 0.105 & 0.085 & 0.04 & 0.04 & 0.07 \\
800 & 0.05 & 0.005 & 0.055 & 0.075 & 0.03 & 0.035 & 0.055 \\
1000 & 0.075 & 0.005 & 0.045 & 0.045 & 0.015 & 0.02 & 0.05 \\
\bottomrule
\end{tabular}
\end{small}
\end{center}
\end{table}

\begin{table}[h!]%
\caption{Two-sample level experiment using uniform densities varying the dimension.}%
\label{tab:mmdd}%
\vspace{-0.6cm}%
\begin{center}%
\begin{small}%
\begin{tabular}{cccccccc}%
\toprule%
Dimension%
& ME% 
& SCF% 
& \begin{tabular}{@{}c@{}}OST \\ PSI\end{tabular}%  
& \begin{tabular}{@{}c@{}}MMDAggInc \\ $R=1$\end{tabular}%
& \begin{tabular}{@{}c@{}}MMDAggInc \\ $R=100$\end{tabular}%
& \begin{tabular}{@{}c@{}}MMDAggInc \\ $R=200$\end{tabular}%
& MMDAggCom%
\\%
\midrule%
1 & 0.045 & 0     & 0.035 & 0.02  & 0.045 & 0.04  & 0.045 \\
2 & 0.045 & 0.035 & 0.085 & 0.1   & 0.05  & 0.04  & 0.035 \\
3 & 0.04  & 0.05  & 0.04  & 0.04  & 0.05  & 0.06  & 0.025 \\
4 & 0.045 & 0.05  & 0.03  & 0.055 & 0.045 & 0.045 & 0.03 \\
\bottomrule
\end{tabular}
\end{small}
\end{center}
\end{table}

\begin{table}[h!]
\caption{Independence level experiment using uniform densities varying the sample size.}
\label{tab:hsicn}
\vspace{-0.3cm}
\begin{center}
\begin{small}
\begin{tabular}{cccccc}
\toprule
\begin{tabular}{@{}c@{}}Sample \\ size\end{tabular} 
& FSIC
& \begin{tabular}{@{}c@{}}HSICAggInc \\ $R=1$\end{tabular}  
& \begin{tabular}{@{}c@{}}HSICAggInc \\ $R=100$\end{tabular}
& \begin{tabular}{@{}c@{}}HSICAggInc \\ $R=200$\end{tabular}
& HSICAggCom
\\
\midrule
200  & 0.04  & 0.055 & 0.035 & 0.035 & 0.035 \\
400  & 0.045 & 0.05  & 0.04  & 0.05  & 0.05  \\
600  & 0.05  & 0.035 & 0.05  & 0.06  & 0.05  \\
800  & 0.03  & 0.07  & 0.02  & 0.035 & 0.04  \\
1000 & 0.07  & 0.02  & 0.085 & 0.035 & 0.04  \\
\bottomrule
\end{tabular}
\end{small}
\end{center}
\end{table}

\begin{table}[h!]
\caption{Independence level experiment using uniform densities varying the dimension.}
\label{tab:hsicd}
\vspace{-0.3cm}
\begin{center}
\begin{small}
\begin{tabular}{cccccc}
\toprule
Dimension
& FSIC
& \begin{tabular}{@{}c@{}}HSICAggInc \\ $R=1$\end{tabular}  
& \begin{tabular}{@{}c@{}}HSICAggInc \\ $R=100$\end{tabular}
& \begin{tabular}{@{}c@{}}HSICAggInc \\ $R=200$\end{tabular}
& HSICAggCom
\\
\midrule
2 & 0.035 & 0.065 & 0.08  & 0.055 & 0.07  \\
3 & 0.065 & 0.055 & 0.035 & 0.02  & 0.025 \\
4 & 0.04  & 0.035 & 0.045 & 0.055 & 0.055 \\
\bottomrule
\end{tabular}
\end{small}
\end{center}
\end{table}

\begin{table}[h!]
\caption{Goodness-of-fit level experiment using a Gaussian-Bernoulli Restricted Boltzmann Machine varying the sample size.}
\label{tab:ksdn}
\vspace{-0.3cm}
\begin{center}
\begin{small}
\begin{tabular}{ccccccc}
\toprule
\begin{tabular}{@{}c@{}}Sample \\ size\end{tabular} 
& FSSD
& LSD
& \begin{tabular}{@{}c@{}}KSDAggInc \\ $R=1$\end{tabular}  
& \begin{tabular}{@{}c@{}}KSDAggInc \\ $R=100$\end{tabular}
& \begin{tabular}{@{}c@{}}KSDAggInc \\ $R=200$\end{tabular}
& KSDAggCom
\\
\midrule
200  & 0.02  & 0.07 & 0.05  & 0.045 & 0.06  & 0.06  \\
400  & 0.03  & 0.04 & 0.06  & 0.04  & 0.065 & 0.055  \\
600  & 0.04  & 0.075 & 0.03  & 0.03  & 0.04  & 0.07  \\
800  & 0.03  & 0.06 & 0.055 & 0.06  & 0.045 & 0.07   \\
1000 & 0.025 & 0.05 & 0.045 & 0.035 & 0.045 & 0.065  \\
\bottomrule
\end{tabular}
\end{small}
\end{center}
\end{table}

\begin{table}[H]
\caption{Goodness-of-fit level experiment using a Gaussian-Bernoulli Restricted Boltzmann Machine varying the dimension.}
\label{tab:ksdd}
\vspace{-0.3cm}
\begin{center}
\begin{small}
\begin{tabular}{ccccccc}
\toprule
Dimension
& FSSD
& LSD
& \begin{tabular}{@{}c@{}}KSDAggInc \\ $R=1$\end{tabular}  
& \begin{tabular}{@{}c@{}}KSDAggInc \\ $R=100$\end{tabular}
& \begin{tabular}{@{}c@{}}KSDAggInc \\ $R=200$\end{tabular}
& KSDAggCom
\\
\midrule
20 & 0.02  & 0.055 & 0.045 & 0.06  & 0.065 & 0.05 \\
40 & 0.04  & 0.055 & 0.07  & 0.055 & 0.065 & 0.07 \\
60 & 0.04  & 0.055 & 0.06  & 0.04  & 0.05  & 0.06 \\
80 & 0.015 & 0.04 & 0.045 & 0.04  & 0.035 & 0.05  \\
\bottomrule
\end{tabular}
\end{small}
\end{center}
\end{table}

\subsection{Aggregation experiments}

\begin{figure}[h!]
  \centering
  \includegraphics[width=\textwidth]{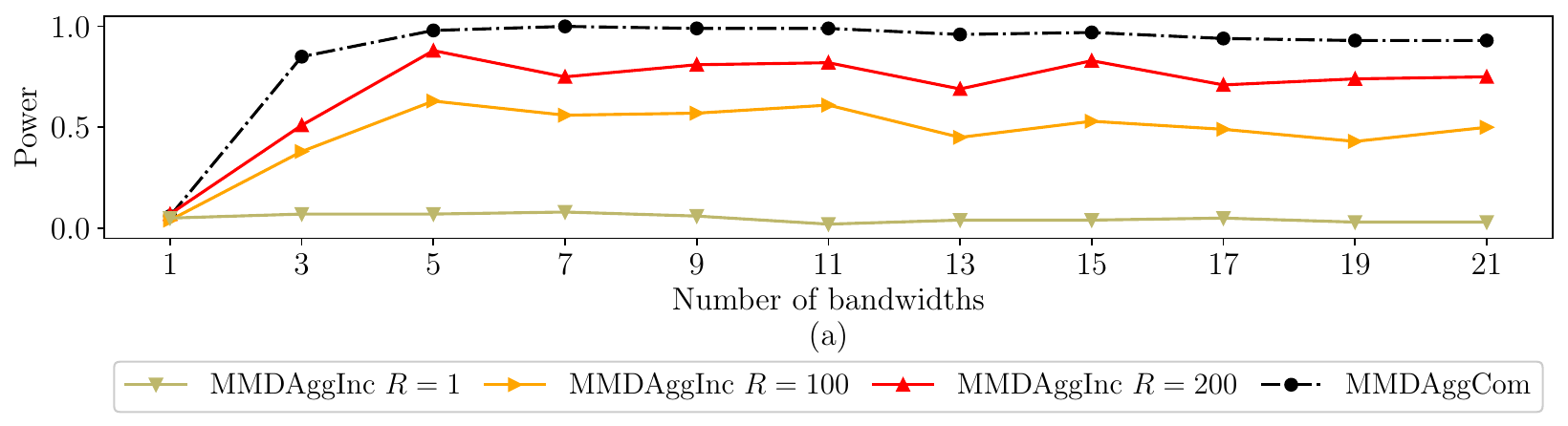}
  \includegraphics[width=\textwidth]{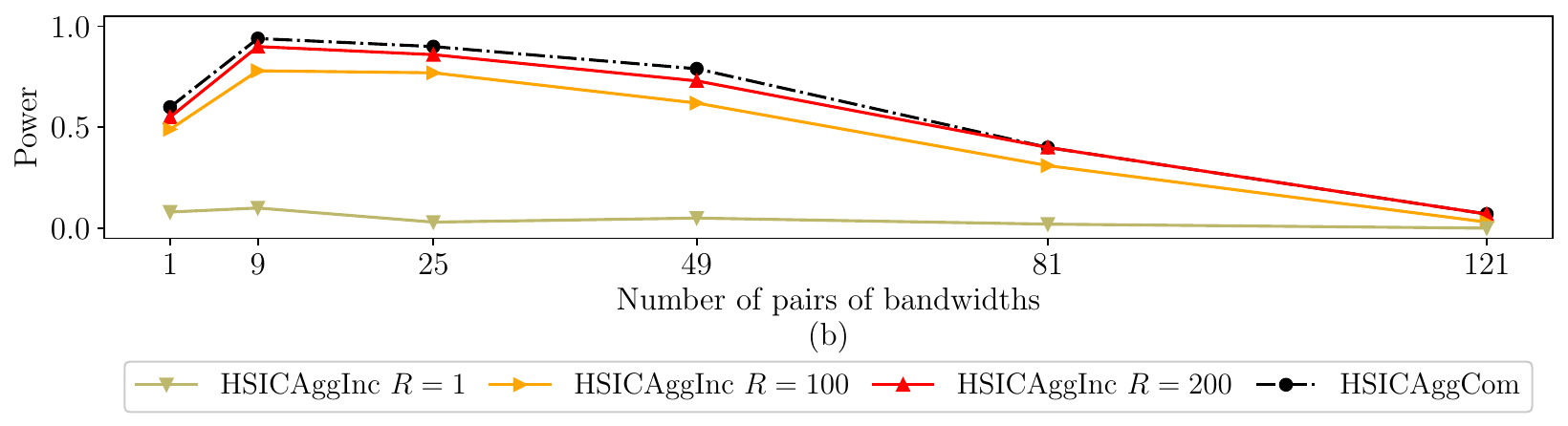}
  \includegraphics[width=\textwidth]{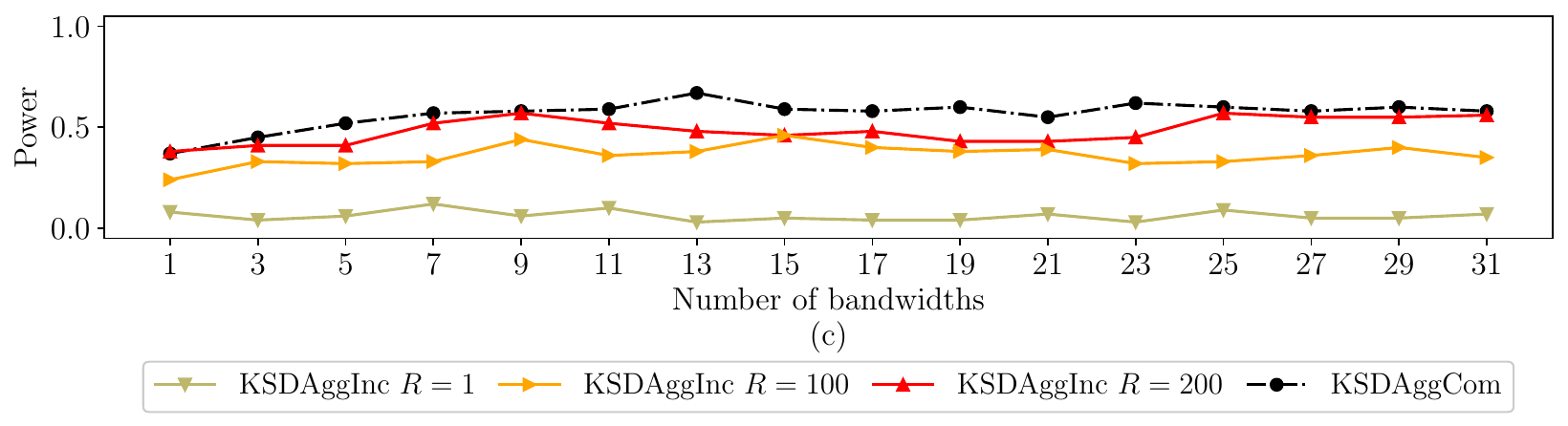}
  \caption{%
  Increasing the collection of bandwidths in the experimental setting of \Cref{fig:experiments} for the {\em (a)}~two-sample , {\em (b)} independence, and {\em (c)} goodness-of-fit problems. 
  The power results are averaged over 100 repetitions.
  \label{fig:experiments_col}
  }
\end{figure}

We illustrate in \Cref{fig:experiments_col} the benefits of the aggregation procedure by starting from a `collection' consisting of only the median bandwidth and increasing the size of the collection by adding more bandwidths. 
In all three settings, we observe that the power for the test with only the median bandwidth  is low. 
As we increase the number of bandwidths, the power first increases as the test has access to `better-suited' bandwidths. 

For MMDAggInc and KSDAggInc, once the optimal bandwidth is included in the collection, the power reaches a plateau. 
We do not pay a price in power for considering more bandwidths (or kernels), and so the user is encouraged to consider many kernels with various bandwidths.
For the unscaled Gaussian kernel, we are essentially aggregating over kernel matrices which interpolate between the identity matrix (as the bandwidth goes to $0$) and the matrix of ones (as the bandwidth goes to $\infty$).

The HSICAggInc case is more challenging: since there are pairs of kernels, the total number of bandwidth combinations grows rapidly (\emph{e.g.} $b$ bandwidths for each kernel corresponds to $b^2$ pairs of kernels). 
In this case, we observe a significant decay in test power once more than $49$ bandwidths are considered.

\FloatBarrier

\subsection{Parameter choice for L1 IMQ and Cauchy RFF tests of \citet{huggins2018random} }
\label{sec:ndraws}

As in the experiments section of \citet{huggins2018random} (and as for FSSD), $10$ features are used when running Cauchy RFF and L1 IMQ. 
In their \href{https://bitbucket.org/jhhuggins/random-feature-stein-discrepancies/}{implementation} for their experiments, they draw $5000$ samples from the unnormalized density for covariance matrix estimation to simulate the null hypothesis (code: \texttt{RFDH0SimCovDrawV(n\_draw=5000})). 
This procedure causes long runtimes of roughly 16 seconds; this is much more computationally expensive than simulating the null using a wild bootstrap as KSDAggInc does.

We tried different values for {\texttt{n\_draw} and found that using $\texttt{n\_draw}=500$ has almost no effect on the test power and reduces the runtimes from 16 seconds for $\texttt{n\_draw}=5000$, to 2 seconds (as reported in \Cref{fig:experiments} (l)) for $\texttt{n\_draw}=500$. 
We tried smaller values than $500$ for \texttt{n\_draw} but this drastically decreased test power.
We also verified that the test still has well-calibrated level when using $\texttt{n\_draw}=500$. 
We have used this tuned parameter $\texttt{n\_draw}=500$ in our experiments in \Cref{fig:experiments} (i--l).
In \Cref{fig:comparison}, we show the power and runtime differences when using 500 or 5000 for \texttt{n\_draw} for L1 IMQ and Cauchy RFF in the setting considered in \Cref{fig:experiments} (i--l), we also plot the test power and runtimes achieved by KSDAggInc $R=200$.

\begin{figure}[h!]
    \centering
    \includegraphics[width=\textwidth]{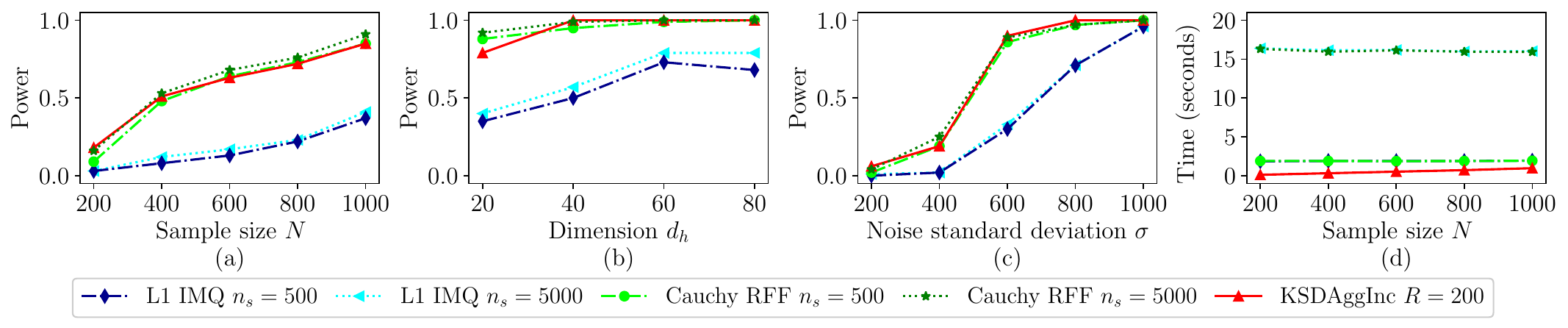}
    \caption{%
    Goodness-of-fit experiment using a Gaussian-Bernoulli Restricted Boltzmann Machine.
    We consider L1 IMQ and Cauchy RFF tests of \citet{huggins2018random} drawing either $n_s=\texttt{n\_draw}=5000$ or $n_s=\texttt{n\_draw}=500$ samples to simulate the null hypothesis. 
    The power results are averaged over 100 repetitions and the runtimes over 20 repetitions. 
    }
    \label{fig:comparison}
\end{figure}

\FloatBarrier

\section{Discussions}
In this section, we provide detailed discussions on several subjects. 
We present the motivation behind the definitions of the MMD and HSIC estimators of \Cref{Uincmmd,Uinchsic}. 
We also explain how to define a different incomplete MMD $U$-statistic which is better-suited to the case of unbalanced sample sizes, we point out the challenges arising from working with this estimator.
Finally, we provide details on comparison with related work, and on future research directions.

\subsection{Motivation behind expressions~\eqref{Uincmmd} and \eqref{Uinchsic}}
\label{sec:hsicincref}

For the two-sample problem, Equation (26) of \citet[][Section 6.1]{kim2020minimax} gives an expression of the MMD $U$-statistic as
\begin{align*}
	U_1 = \frac{1}{\abss{\textbf{i}_2^m}\abss{\textbf{i}_2^n}} 
	\sum_{(i,i')\in \textbf{i}_2^m}
	\sum_{(j,j')\in \textbf{i}_2^n}
	h_k^{\mathrm{MMD}}(X_i, X_{i'}; Y_j, Y_{j'}).
\end{align*}
Now, one way to construct an incomplete MMD $U$-statistic would be to replace those two complete sums above with two incomplete sums (see Appendix~\ref{Section: unbalanced sample size}), but we do not want to take this approach in order to keep a unified framework across the three testing frameworks. We instead take the summation over $(i,i')\in \textbf{i}_2^N$ and obtain an estimator
\begin{align*}
	U_2 = \frac{1}{\abss{\textbf{i}_2^{N}}}
	\sum_{(i,i')\in \textbf{i}_2^N}
	h_k^{\mathrm{MMD}}(X_{i},X_{i'};Y_{i},Y_{i'}),
\end{align*}
where $N = \min(m,n)$. By assuming $N = m$, we denote by $\{L_1,\ldots,L_n\}$ a random permutation of $\{1,\ldots,n\}$. As noted by \citet[][Section 6.1]{kim2020minimax}, the expectation of 
\begin{align*}
	\frac{1}{\abss{\textbf{i}_2^{N}}}
	\sum_{(i,i')\in \textbf{i}_2^N}
	h_k^{\mathrm{MMD}}(X_{i},X_{i'};Y_{L_i},Y_{L_{i'}})
\end{align*}
over $\{L_1,\ldots,L_n\}$ is equal to $U_1$. This motivates our choice of incomplete MMD estimator in \Cref{Uincmmd} of our paper, which can be regarded as a generalization of $U_2$ above. Similarly, as discussed in \citet[][Section 6.2]{kim2020minimax}, the complete HSIC $U$-statistic in \Cref{Uhsic} can be viewed as the average of incomplete $U$-statistics. More specifically, let $\{L_1,\ldots,L_{\lfloor N/2 \rfloor}\}$ be a $\lfloor N/2 \rfloor$-tuple uniformly sampled without replacement from $\{1,\ldots,N\}$, and let $\{\tilde{L}_1,\ldots,\tilde{L}_{\lfloor N/2 \rfloor}\}$ be another $\lfloor N/2 \rfloor$-tuple uniformly sampled without replacement from $\{1,\ldots,N\} \setminus \{L_1,\ldots,L_{\lfloor N/2 \rfloor}\}$. Then, the $U$-statistic in \Cref{Uhsic} is the expectation of  
\begin{align*}
	\frac{1}{\abss{\textbf{i}_2^{\lfloor N/2\rfloor}}}
	\sum_{(i_1,i_2)\in \textbf{i}_2^{\lfloor N/2\rfloor}}
	h^{\mathrm{HSIC}}_{k,\ell}(Z_{L_{i_1}},Z_{L_{i_2}};Z_{\tilde{L}_{i_1}},Z_{\tilde{L}_{i_2}}),
\end{align*}
over $\{L_1,\ldots,L_{\lfloor N/2 \rfloor}, \tilde{L}_1,\ldots,\tilde{L}_{\lfloor N/2 \rfloor}\}$. This motivates the definition of our incomplete HSIC estimator in \Cref{Uinchsic}.

\subsection{Incomplete MMD $U$-statistic with unbalanced sample sizes} \label{Section: unbalanced sample size}
Our incomplete $U$-statistic of \Cref{Uincmmd} for the two-sample problem is constructed using the minimum between $m$ and $n$.
If the sample sizes are of the same order of magnitude, then this is not restrictive since we are interested in using only a subset of entries of the kernel matrix in the first place.
However, in the setting in which the difference between $m$ and $n$ is of several orders of magnitude, our estimator in \Cref{Uincmmd}  does not effectively incorporate the unbalanced sample sizes. 
When the sample sizes are highly unbalanced, one could instead consider an alternative incomplete $U$-statistic given as 
\begin{align*}
	U_{\mathrm{imb}} = & \frac{1}{|\mathcal{D}_m| |\mathcal{D}_n|}
	\sum_{(i,j)\in \mathcal{D}_m}
	\sum_{(r,s)\in \mathcal{D}_n}
	h_k^{\mathrm{MMD}}(X_i, X_j; Y_r, Y_s) \\
	= & \frac{1}{|\mathcal{D}_m| |\mathcal{D}_n|}
	\sum_{(i,j)\in \mathcal{D}_m}
	\sum_{(r,s)\in \mathcal{D}_n}
	\Big(k(X_i,X_j)  - k(X_i,Y_s) - k(X_j,Y_r) + k(Y_r,Y_s)\Big).	
\end{align*}
This expression, for example, results in a linear-time test for the choices $|\mathcal{D}_m| = c \sqrt{m}$ and $|\mathcal{D}_n| = c' \sqrt{n}$ for positive constants $c$ and $c'$ since $|\mathcal{D}_m| |\mathcal{D}_n| = c c' \sqrt{m} \sqrt{n} \leq c c' \mathrm{max}(m,n)$. Other choices of design sizes are also possible to obtain linear-time tests. While this estimator is natural for the unbalanced scenario, the form of the test statistic does not allow us to use a wild bootstrap. Instead, one may need to rely on the permutation procedure to calibrate the test statistic, which leads to several theoretical and practical challenges explained below.

\textbf{Theory.} From a theoretical side, it is possible to derive a variance bound (corresponding to \Cref{lem:varbound}) for the alternative estimator $U_{\mathrm{imb}}$. However, deriving a quantile bound (corresponding to \Cref{lem:Uincbound}) for a permuted version of $U_{\mathrm{imb}}$ is highly non-trivial: the extension of the result of \citet[][Theorem 6.3]{kim2020minimax} to the case of the permuted version of $U_{\mathrm{imb}}$ is ongoing work.

\textbf{Practice.} Theoretically, the cost of computing $B$ permuted estimates is $\mathcal{O}(B|\mathcal{D}_m| |\mathcal{D}_n|)$ which would be the same as if we could use a wild bootstrap. However, in practice, the computational time will be much higher because for each permuted estimate, we need to evaluate the kernel matrix at new permuted pairs (possibly outside of the original design), while for the wild bootstrap we do not need to compute any extra kernel values: this changes the computation times drastically. In order to avoid this, we would need to restrict ourselves to permutations for which we have already computed kernel values using the fact that 
$h_k^{\mathrm{MMD}}(X_i, Y_s; Y_r,X_j) = - h_k^{\mathrm{MMD}}(X_i, X_j; Y_r, Y_s)$. It remains as future work to study conditions under which the set of such permutations is larger than the set consisting of the identity only, and is also large enough to construct accurate quantiles.

\subsection{Comparison with \cite{li2019optimality}}
\cite{li2019optimality} also consider the three testing problems and study minimax optimality/adaptivity of their procedures over Sobolev balls. We now discuss and differentiate the approach by \cite{li2019optimality} from ours. First of all, their tests run in quadratic time and control the probability of type I error only asymptotically, while our proposed tests have well-calibrated non-asymptotic levels over a broader class of null distributions and are computationally efficient. Their theoretical guarantees hold only for the Gaussian kernel and with the smoothness restriction that $s>d/4$ while ours hold for a wide range of kernels (see \Cref{kernels}) and for any $s>0$ (see \Cref{theo:agg}). 
We also point out that while they assume that both densities $p$ and $q$ lie in a Sobolev ball, we only require that their difference $p-q$ belongs to the Sobolev ball.
Note that, they tackle the goodness-of-fit problem in a different way. They do not use the KSD and instead use a one-sample MMD with some expectations of the Gaussian kernel under the model. For a generic model density, one cannot compute such an expectation explicitly and hence cannot use their proposed test. In contrast, the KSD that we consider does not suffer from the same issue and it is more broadly applicable. 

\subsection{Potential future research directions}
This subsection discusses potential directions for future work. 

\textbf{Interpretable tests.}
When the aggregated test rejects the null hypothesis, the test returns all kernels whose associated single test with adjusted level has rejected the null. 
We stress that this is done using all of the samples, without resorting to data splitting.
Those kernels returned by the test are the ones which are well-suited to detect the difference in densities.
They can therefore be analysed and interpreted to obtain some information which can help the user understand how the densities differ from each other.
For example, we could observe that the densities differ at some specific lengthscales, from which we can infer whether the distribution shift is local, global, or both. 
If the kernels use different features, we can get a better understanding of the type of features which capture best the difference in densities. 
This interpretability of the results of our AggInc tests could be very useful, we will further explore it in upcoming work.

\textbf{Beyond linear time tests.} Potential directions for future work include studying the regime with $L\lesssim N$, which corresponds to `faster than linear' tests. For this sub-linear case, our results do not give a definite answer to the question as to whether the separation rate converges to zero. Future work would focus on either deriving tighter bounds that converge to zero in this regime, or proving that the uniform separation rate is bounded below in this setting.

\textbf{Computational and statistical trade-off.} As shown in \Cref{theo:fixed}, the quadratic-time MMD and HSIC tests are minimax rate optimal over Sobolev balls. To the best of our knowledge, it is unknown whether there exists a sub-quadratic time test that achieves the same rate optimality. Indeed, the current literature is mostly silent on optimising the power under computational constraints. \Cref{theo:fixed}~(ii) demonstrates a trade-off between the computational budget and the separation rate focusing on incomplete $U$-statistics, but our result does not tell us whether this trade-off is (universally) tight. We think this is one of the limitations of our work and hope that a follow-up study can make progress on this topic.

\textbf{Continuously optimising a kernel without data splitting.} In order to achieve a competitive power performance over a large class of alternatives, we combine finitely many kernels and construct an aggregated test. 
There is another line of work that considers a continuous collection of kernels (for example, indexed by the bandwidth parameter on the positive line) and chooses the kernel that maximises the (empirical) power. 
As far as we are aware, the current approach to continuously optimizing a kernel relies on data splitting, which often negatively affects the power performance.
While our aggregated tests do not require data splitting and can retain high power even for large collections of kernels, in certain scenarios there might exist some important range of kernels not included in our finite collection which would lead to more powerful tests.
It is therefore interesting to further extend our approach to the case of a continuous collection of kernels and investigate its statistical properties.
We note that, for MMDAggInc and KSDAggInc, as also observed by \citet{schrab2021mmd,schrab2022ksd}, we find empirically that  increasing the number of bandwidths does not result in a loss of power, and corresponds to using a finer discretisation of the intervals considered. The continuous case is the limit of the discretisation.

\textbf{Improving conditions.} Lastly, it would be interesting to see if the polynomial factor of $\beta$ in our power guaranteeing conditions of \Cref{theo:fixed,theo:agg} can be improved using a sharper concentration bound. Also, future work can be dedicated to figuring out whether the dependence of $\ln (\ln (N))$ in the adaptive rate can be improved. We leave these important and interesting questions to future work.

\section{Proofs}

\subsection{Proof of \Cref{prop:level}}
\label{proof:level}

The asymptotic level of the goodness-of-fit test using a wild bootstrap follows from the results of \citet{shao2010dependent}, \citet{leucht2013dependent}, \citet{chwialkowski2014wild,chwialkowski2016kernel}.
As pointed out by \citet{schrab2022ksd} for the complete $U$-statistics, the KSD test statistic and the wild bootstrapped KSD statistics are not exchangeable under the null, and hence non-asymptotic level cannot be proved using the result of \citet[Lemma 1]{romano2005exact}.

The non-asymptotic level for the two-sample test follows exactly from the reasoning of \citet[Propositon 1]{schrab2021mmd}.
The fact that we work with incomplete $U$-statistics rather than with their complete counterparts does not affect the proof of exchangeability of $U_\lambda^1,\dots,U_\lambda^{B_1+1}$.

For the independence problem, \citet[Proposition 1]{albert2019adaptive} prove that the quadratic-time HSIC estimator and the permuted test statistics are exchangeable under the null hypothesis, it remains to be shown that this also holds in our incomplete setting using a wild bootstrap. 
Assuming that exchangeability under the null holds, the desired non-asymptotic level $\alpha$ can then be guaranteed using the result of \citet[Lemma 1]{romano2005exact}, exactly as done by \citet[Proposition~1]{albert2019adaptive}.

We now prove that $U_\lambda^1,\dots,U_\lambda^{B_1+1}$ for the independence problem are exchangeable under the null. Since $U_\lambda^1,\dots,U_\lambda^{B_1}$ are i.i.d.\ given the data, they are exchangeable under the null. So, we need to prove that 
\begin{equation}
\label{prop1eq1}
\sum_{(i,j)\in \D_{\lfloor N/2\rfloor}}
h_{k,\ell}^{\mathrm{HSIC}}\left(Z_i, Z_j, Z_{i+\lfloor N/2\rfloor}, Z_{j+\lfloor N/2\rfloor}\right) 
\end{equation}
is, under the null, distributed like
\begin{equation}
\label{prop1eq2}
\sum_{(i,j)\in \D_{\lfloor N/2\rfloor}}
\epsilon_i\epsilon_j
h_{k,\ell}^{\mathrm{HSIC}}\!\left(Z_i, Z_j, Z_{i+\lfloor N/2\rfloor}, Z_{j+\lfloor N/2\rfloor}\right) 
\end{equation}
where $\epsilon_1,\dots,\epsilon_N$ are i.i.d.\ Rademacher random variables.
Using the result of \citet[Appendix B, Proposition 11]{schrab2021mmd}, considering the identity $s_1(i) = i$ for $i=1,\dots,2\lfloor N/2\rfloor$, and the swap function $s_{-1}(i) = i+\lfloor N/2\rfloor$ and $s_{-1}(i+\lfloor N/2\rfloor) = i$ for $i=1,\dots,2\lfloor N/2\rfloor$, we have 
\begin{align*}
&\sum_{(i,j)\in \D_{\lfloor N/2\rfloor}}
\epsilon_i\epsilon_j
h_{k,\ell}^{\mathrm{HSIC}}\!\left(Z_i, Z_j, Z_{i+\lfloor N/2\rfloor}, Z_{j+\lfloor N/2\rfloor}\right) \\
=~ 
&\frac{1}{4}
\sum_{(i,j)\in \D_{\lfloor N/2\rfloor}}
h_k^{\mathrm{MMD}}\!\!\left(X_i, X_j; X_{i+\lfloor N/2\rfloor}, X_{j+\lfloor N/2\rfloor}\right)
\!\Big(
\epsilon_i\epsilon_j h_\ell^{\mathrm{MMD}}\!\left(Y_i, Y_j; Y_{i+\lfloor N/2\rfloor}, Y_{j+\lfloor N/2\rfloor}\right)\!\!
\Big) \\
=~ 
&\frac{1}{4}
\!\!\!\sum_{\substack{(i,j)\in\\ \D_{\lfloor N/2\rfloor}}}
\!\!\!h_k^{\mathrm{MMD}}\!\left(X_i, X_j; X_{i+\lfloor N/2\rfloor}, X_{j+\lfloor N/2\rfloor}\right)
\!h_\ell^{\mathrm{MMD}}\!\big(Y_{s_{\epsilon_i}\!(i)}, Y_{s_{\epsilon_j}\!(j)}; Y_{s_{\epsilon_i}\!(i+\lfloor N/2\rfloor)}, Y_{s_{\epsilon_j}\!(j+\lfloor N/2\rfloor)}\big) \\
=~
&\abss{\D_{\lfloor N/2\rfloor}}\,\overline{\mathrm{HSIC}}_{k,\ell}\big(\Zn^\epsilon; \D_{\lfloor N/2\rfloor} \big) 
\end{align*}
where
$
\Zn^\epsilon \coloneqq \p{
\pbig{X_i, Y_{s_{\epsilon_i}\!(i)}}
}_{1\leq i \leq N}
$
with $Y_{s_{\epsilon_i}\!(i)}\in \{Y_i,Y_{i+\lfloor N/2\rfloor}\}$ for $i=1,\dots,\lfloor N/2\rfloor$ and $Y_{s_{\epsilon_i}\!(i)}\in \{Y_i,Y_{i-\lfloor N/2\rfloor}\}$ for $i =\lfloor N/2\rfloor+1,\dots,2\lfloor N/2\rfloor$.
Now, under the null, the variables $(X_i)_{1\leq i \leq N}$ and $(Y_i)_{1\leq i \leq N}$ are independent, so $\Zn^\epsilon$ is distributed like $\Zn$.
We deduce that $\overline{\mathrm{HSIC}}_{k,\ell}\big(\Zn; \D_{\lfloor N/2\rfloor} \big)$ and $\overline{\mathrm{HSIC}}_{k,\ell}\big(\Zn^\epsilon; \D_{\lfloor N/2\rfloor} \big)$ have the same distribution under the null, and hence, that the terms in \Cref{prop1eq1,prop1eq2} also have the same distribution under the null.
We deduce that $U_\lambda^1,\dots,U_\lambda^{B_1+1}$ for the independence problem are exchangeable under the null, which completes the proof.

\subsection{Proof of \Cref{lem:varbound}}
\label{proof:varbound}

Consider the case of fixed design.
Using the variance expression of \citet[Theorem 2, p.\! 190]{lee1990ustatistic}, we have
$$
\mathrm{var}\!\left(\overline{U}\right)
=
\frac{f_1 \sigma_1^2 + f_2 \sigma_2^2}{\abs{\D}^2}
$$
where $f_i$ is the number of pairs of sets in the design $\D$ that have $i$ elements in common.
The pairs of sets in $\D$ with 2 elements in common are $\left\{\p{\{i,j\},\{i,j\}}: (i,j)\in\D\right\}$, so $f_2=\abs{\D}$.
We now calculate the number of pairs of sets in $\D$ with 1 element in common.
We start with a pair $(i,j)\in\D$ (there are $\abs{\D}$ such pairs).
The number of pairs in $\D$ which have one element in common with $(i,j)$ is upper bounded by the number of pairs in $\mathbf{i}_2^N$ which have one element in common with $(i,j)$, those are
$\{\{i,r\}: 1\leq r\leq N, r\neq i\}\cup \{\{j,r\}: 1\leq r\leq N, r \neq j\}$ of size smaller than $2N$.
We deduce that $f_1\leq 2N\abs{\D}$.
Combining those results, we obtain 
$$
\mathrm{var}\!\left(\overline{U}\right)
\leq
\frac{f_1 \sigma_1^2 + f_2 \sigma_2^2}{\abs{\D}^2}
\leq \frac{2N}{\abs{\D}}\sigma_1^2 + \frac{1}{\abs{\D}}\sigma_2^2
$$
as desired.

Let us now consider the random design case. 
Recall that using the variance expression of the complete $U$-statistic of \citet[Theorem 3 p.\! 12]{lee1990ustatistic}, which we denote $U$, we can obtain that
$$
\mathrm{var}\!\left(U\right) 
\lesssim 
\frac{\sigma_1^2}{N} + \frac{\sigma_2^2}{N^2}
$$
as done by \citet[Appendix E]{kim2020minimax} and \citet[Lemma 10]{albert2019adaptive}.
Using the result of \citet[Theorem 4 p.\! 193]{lee1990ustatistic}, the variance of the incomplete $U$-statistic $\overline{U}$ can be expressed in terms of the variance of the complete $U$-statistic $U$.
For random design with replacement, we have
\begin{align*}
    \mathrm{var}\!\left(\overline{U}\right)
    &= \frac{\sigma_2^2 }{\abs{\D}}
    + \p{1-\frac{1}{\abs{\D}}} \mathrm{var}\!\left(U\right) \\
    &\lesssim
    \frac{\sigma_1^2}{N} + \p{\frac{1}{\abs{\D}} + \frac{1}{N^2}}\sigma_2^2 \\
    &\lesssim
    \frac{\sigma_1^2}{N} + \frac{1}{\abs{\D}} \sigma_2^2.
\end{align*}
Letting $S\coloneqq N(N-1)/2$, for random design without replacement, we have
\begin{align*}
    \mathrm{var}\!\left(\overline{U}\right)
    &= \frac{S- \abs{\D}}{\abs{\D}(S-1)} \sigma_2^2
    + \frac{S}{S-1}\p{1-\frac{1}{\abs{\D}}} \mathrm{var}\!\left(U\right) \\
    &\lesssim 
    \frac{\sigma_1^2}{N} + \p{\frac{1 }{\abs{\D}} + \frac{1}{N^2}}\sigma_2^2 \\
    &\lesssim 
    \frac{\sigma_1^2}{N} + \frac{1 }{\abs{\D}} \sigma_2^2.
\end{align*}
We have used the fact that $\frac{1}{N^2} \leq \frac{1}{\abs{\D}}$ since $\abs{\D}\leq N^2$.

\subsection{Proof of \Cref{lem:Uincbound}}
\label{proof:Uincbound}

We rely on the concentration bound for i.i.d.\ Rademacher chaos of \citet[Corollary 3.2.6]{pena1999decoupling} which as presented in \citet[Equation (39)]{kim2020minimax} takes the form
$$
\PP_\epsilon\p{
    \bigg|
        \sum_{(i,j)\in\mathbf{i}_2^N} \epsilon_i \epsilon_j a_{i,j}
    \bigg|
    \geq
    t
}
\leq 
2
\exp\p{
    -{Ct}{
        \p{\sum_{(i,j)\in\mathbf{i}_2^N} a_{i,j}^2}^{-1}
    }
}
$$
for some constant $C>0$ and for every $t\geq 0$, where $\epsilon_1,\dots,\epsilon_N$ are i.i.d.\ Rademacher random variables taking values in $\{-1,1\}$. Letting 
$$
a_{i,j} \coloneqq \frac{h(Z_i,Z_j)}{\abs{\D}} \one\Big[(i,j)\in\D\Big]
\qquad \textrm{for} \qquad
(i,j)\in\mathbf{i}_2^N,
$$
where $\one$ denotes the indicator function,
we obtain 
\begin{align*}
\PP_\epsilon\p{
    \frac{1}{\abs{\D}}
    \bigg|
        \sum_{(i,j)\in\D} \epsilon_i \epsilon_j h(Z_i,Z_j)
    \bigg|
    \geq
    t
    ~\Big|~ \Zn, \D
}
&\leq 
2
\exp\p{
    -{Ct}{
        \p{\frac{1}{\abs{\D}^2}\sum_{(i,j)\in\D} h(Z_i,Z_j)^2}^{-1}
    }
} \\
&\leq 
2
\exp\p{
    -{Ct}{
        \p{\frac{1}{\abs{\D}^2}\sum_{(i,j)\in\mathbf{i}_2^N} h(Z_i,Z_j)^2}^{-1}
    }
} 
\end{align*}
which concludes the proof.

\subsection{Proof of \Cref{theo:fixed}}
\label{proof:fixed}

We start by reviewing the steps of the proofs of \citet{albert2019adaptive} and \citet{schrab2021mmd,schrab2022ksd} who prove that, for each of the three respective testing frameworks, a sufficient condition to ensure control of the probability of type II error for the quadratic-time test is the existence of a constant $C>0$ such that
\begin{equation}
    \label{eq:conditionquadratic}
    \|{p-q}\|^2_2 ~\geq~ \|{(p-q)-T_\lambda(p-q)}\|^2_2 + C \frac{1}{N}\frac{\ln\!\p{{1}/{\alpha}}}{\beta} \sigma_{2,\lambda}.
\end{equation}
Those quadratic-time tests  use the complete $U$-statistics defined in \Cref{Ummd,Uhsic,Uksd}, which we denote as $U_\lambda$.
The key results for their proofs rely on deriving variance and quantile bounds.

The variance bound is of the form
\begin{equation}
    \label{eq:varboundquad}
    \mathrm{var}\!\left(U_\lambda\right)
    \lesssim
    \frac{1}{N}\sigma_1^2 + \frac{1}{N^2}\sigma_2^2
\end{equation}
where they show that, for $h_\lambda\in\big\{h_{k_\lambda}^{\mathrm{MMD}},h_{k_\lambda,\ell_\mu}^{\mathrm{HSIC}},h_{k_\lambda,p}^{\mathrm{KSD}}\big\}$ defined in \Cref{h_mmd,h_hsic,h_ksd}, we have
$$
\sigma_{1,\lambda}^2 \coloneqq \mathrm{var}\!\p{\mathbb{E}\!\left[h_\lambda(Z,Z')\big| Z'\right]}
\lesssim \big\|T_\lambda(p-q)\big\|_2^2
$$
and
\begin{equation}
\label{eq:sigma2}
\sigma_{2,\lambda}^2 
\coloneqq \mathrm{var}\!\p{h_\lambda(Z,Z')}
= \E\!\left[h_\lambda(Z,Z')^2\right]
\lesssim \frac{1}{\LL}
\end{equation}
where the last inequality holds only for $h_\lambda\in\big\{h_{k_\lambda}^{\mathrm{MMD}},h_{k_\lambda,\ell_\mu}^{\mathrm{HSIC}}\big\}$.

The quantile bound \citep[Proposition 4]{schrab2021mmd} is of the form
\begin{equation*}
    \PP\!\p{
    \widehat q_{1-\alpha}^{\,\lambda, U, B_1} \lesssim \frac{1}{N}\frac{1}{\sqrt{\delta}}{\ln\!\p{\frac{1}{\alpha}}}\sigma_{2,\lambda}
    }\geq 1-\delta
\end{equation*}
for $\delta\in(0,1)$, where $\widehat q_{1-\alpha}^{\,\lambda, U, B_1}$ is the quantile obtained using $B_1$ wild bootstrapped similarly to the one defined in \Cref{eq:bootstrapquantile} but using the complete $U$-statistic.
Relying on Dvoretzky–Kiefer–Wolfowitz inequality \citep{dvoretzky1956asymptotic,massart1990tight}, \citet[Proposition 4]{schrab2021mmd} show that it suffices to prove the bound 
\begin{equation}
    \label{eq:quantileboundquad}
    \PP\!\p{
    \widehat q_{1-\alpha}^{\,\lambda, U, \infty} \lesssim \frac{1}{N}\frac{1}{\sqrt{\delta}}{\ln\!\p{\frac{1}{\alpha}}}\sigma_{2,\lambda}
    }\geq 1-\delta
\end{equation}
for the true wild bootstrap quantile $\widehat q_{1-\alpha}^{\,\lambda, U, \infty}$ without finite approximation.

Combining those variance and quantile bounds using Chebyshev's inequality \citep{chebyshev1899oeuvres}, they obtain a condition guaranteeing power in terms of the MMD, HSIC and KSD.
By expressing these three measures as an RHKS inner product
$$
\big\langle p-q, T_\lambda(p-q)\big\rangle
= 
\frac{1}{2} \pBig{\|p-q\|_2^2+\|T_\lambda(p-q)\|_2^2 - \|(p-q)-T_\lambda(p-q)\|_2^2},
$$
they obtain the condition in \Cref{eq:conditionquadratic} which guarantees high power in terms of $\|p-q\|_2^2$.
\citet{albert2019adaptive} and \citet{schrab2021mmd} then derive the minimax rate $N^{-2s/(4s+d)}$ over the Sobolev ball $\Sb$ for the independence and two-sample tests using the bandwidths $\lambda_i^* \coloneqq N^{-2/(4s+d)}$ for $i=1,\dots,d$. 

For \Cref{theo:fixed}, we need to obtain the condition in \Cref{eq:conditionquadratic}  with $N$ replaced by $L/N$. 
Hence, following their reasoning, in order to prove \Cref{theo:fixed} (i) \& (ii), it suffices to derive variance and quantiles bounds for incomplete $U$-statistics which have the form of \Cref{eq:varboundquad,eq:quantileboundquad} with $N$ replaced by $L/N$, which we now do.

Using the variance bound for incomplete $U$-statistics $\overline{U}_\lambda$ of \Cref{lem:varbound}, together with the fact that the design size $L\coloneqq\abs{\D}$ is smaller than $N^2$ so that $1/L = L/L^2 \leq N^2/L^2$, we obtain for fixed design that
$$
    \mathrm{var}\!\left(\overline{U}_\lambda\right)
    \lesssim 
        \frac{N}{L}\sigma_{1,\lambda}^2 + \frac{1}{L}\sigma_{2,\lambda}^2
    \lesssim
        \frac{N}{L}\sigma_{1,\lambda}^2 + \p{\frac{N}{L}}^2\sigma_{2,\lambda}^2
    .
$$
We get the same bound for random design since
$$
    \mathrm{var}\!\left(\overline{U}_\lambda\right)
    \lesssim
        \frac{1}{N}\sigma_{1,\lambda}^2 + \frac{1}{L} \sigma_{2,\lambda}^2
    \lesssim
        \frac{N}{L}\sigma_{1,\lambda}^2 + \p{\frac{N}{L}}^2\sigma_{2,\lambda}^2
    ,
$$
as desired.

For the quantile bound, we use \Cref{lem:Uincbound} which, for
$
    A^2_\lambda
    \coloneqq 
    L^{-2}
    \sum_{(i,j)\in \mathbf{i}_{2}^{N}}
    h_\lambda(Z_i,Z_j)^2 
$, gives that there exists some\footnote{For simplicity of notation, we work with $C^{-1}>0$ rather than with $C>0$.} $C>0$ such that
\begin{equation*}
    \PP_\epsilon\!\left({\overline{U}_\lambda^{\epsilon}} \geq t \,\big|\, \Zn, \D\right)
    \leq\PP_\epsilon\!\left(\abss{\overline{U}_\lambda^{\epsilon}} \geq t \,\big|\, \Zn, \D\right)
    \leq 
    2\exp\!\p{
    -\frac{t}{CA_\lambda}
    }.
\end{equation*}
Setting $\alpha\coloneqq 2\exp\!\p{-t/CA_\lambda}$, we obtain
$$
\widehat q_{1-\alpha}^{\,\lambda, \overline{U}, \infty}
= t
= CA_\lambda\ln\pp{2/\alpha}.
$$
For $\delta\in(0,1)$, using Markov's inequality, we obtain 
\begin{equation*}
    \PP\!\p{
    A_\lambda^2 \leq \frac{1}{\delta} \,\E\!\left[A^2_\lambda\right]
    }\geq 1-\delta
\end{equation*}
where
$$
    \E\!\left[A_\lambda^2\right] = \E\!\left[
    \frac{1}{L^2}
    \sum_{(i,j)\in \mathbf{i}_{2}^{N}}
    h(Z_i,Z_j)^2
    \right]
    =
    \frac{N(N-1)}{L^2}\E\!\left[h_\lambda(Z,Z')^2\right]
    \lesssim \frac{N^2}{L^2} \sigma_{2,\lambda}^2
$$
using \Cref{eq:sigma2}. We deduce that
\begin{align*}
    1-\delta
    &\leq
    \PP\!\p{
    A_\lambda^2 \leq \frac{1}{\delta} \,\E\!\left[A^2_\lambda\right]
    }\\
    &=
    \PP\!\p{
    \widehat q_{1-\alpha}^{\,\lambda, \overline{U}, \infty}
    \leq C \frac{1}{\sqrt{\delta}} \ln\!\p{\frac{2}{\alpha}} \sqrt{\E\!\left[A_\lambda^2\right]} }\\
    &\leq 
    \PP\!\p{
    \widehat q_{1-\alpha}^{\,\lambda, \overline{U}, \infty}
    \lesssim \frac{1}{\sqrt{\delta}} \frac{N}{L} \ln\!\p{\frac{1}{\alpha}} \sigma_{2,\lambda}}
\end{align*}
where we absorbed the constant $C$ in the notation `$\lesssim$', and where we used the fact that $\ln(2/\alpha) \lesssim \ln(1/\alpha)$ since $\alpha\in(0,e^{-1})$.
This concludes the proof.

\subsection{Proof of \Cref{theo:indminimax}}
\label{proof:indminimax}

\subsubsection{Proof of  \Cref{theo:indminimax} (i)}
\label{subsec1:indminimax}

In this setting, we consider as proposed by \citet[Equation 32]{kim2020minimax} the permuted HSIC complete $U$-statistic
$$
U^\pi_N \coloneqq \frac{1}{\abs{\mathbf{i}^N_4}} \sum_{(i,j,r,s)\in\mathbf{i}^N_4}
h_{k,\ell}^{\mathrm{HSIC}}\p{
    (X_i,Y_{\pi_i}),
    (X_j,Y_{\pi_j}),
    (X_r,Y_{\pi_r}),
    (X_s,Y_{\pi_s})
}
$$
for a permutation $\pi$ of the indices $\{1,\dots,N\}$, and for $h_{k,\ell}^{\mathrm{HSIC}}$ as defined in \Cref{h_hsic}.

Applying the exponential concentration bound of \citet[Theorem 6.3]{kim2020minimax}, which uses the result of \citet[Theorem 4.1.12]{pena1999decoupling}, we obtain that there exist constants $C_1,C_2>0$ such that
\begin{equation}
    \label{eq:quantileperm}
    \mathbb{P}_\pi (U_N^\pi \geq t ~|~ \Zn)
    \leq 
    C_1 \exp \pp{
    -C_2 \min\pp{
    \frac{Nt}{\Lambda_N}, 
    \frac{Nt^{2/3}}{M_N^{{2/3}}}
    }
    }
\end{equation}
where
$$
\Lambda_N^2 \coloneqq \frac{1}{N^4}
\sum_{i=1}^N
\sum_{j=1}^N
\sum_{r=1}^N
\sum_{s=1}^N
k_\lambda(X_i,X_j)^2\ell_\mu(Y_r,Y_s)^2
$$
and
\begin{align*}
    M_N &\coloneqq \max_{1\leq i,j,r,s \leq N} \big|k_\lambda(X_i,X_j)\ell_\mu(Y_r,Y_s)\big| \\
    &= \max_{1\leq i,j,r,s \leq N} \left|
    \prod_{a=1}^{d_x} \frac{1}{\lambda_a}K_a\pp{\frac{(X_i)_a-(X_j)_a}{\lambda_a}}
    \prod_{b=1}^{d_y} \frac{1}{\lambda_b}L_b\pp{\frac{(Y_r)_b-(Y_s)_b}{\lambda_b}}
    \right| \\
    &\lesssim \frac{1}{\lambda_1\dots\lambda_{d_x}\mu_1\dots\mu_{d_y}}\\
    &= \frac{1}{\LL}
\end{align*}
since the functions $K_1,\dots,K_{d_x}$ and $L_1,\dots,L_{d_y}$ are bounded, and where we recall our notational convention that $d\coloneqq d_x+d_y$ and $\lambda_{d_x+i}\coloneqq \mu_i$ for $i=1,\dots,d_y$.

Using the reasoning of \citet[Proposition 3]{schrab2021mmd}, we see that the results of \citet[Equations C.17, C.18 \& C.19]{albert2019adaptive} hold not only for the Gaussian kernel but more generally for any kernels of the form of \Cref{kernels}.
Those results give us that
\begin{align*}
\E\!\left[k_\lambda(X_1,X_2)^2\ell_\mu(Y_1,Y_2)^2\right] &\lesssim \frac{1}{\lambda_1\dots\lambda_{d_x}\mu_1\dots\mu_{d_y}} = \frac{1}{\LL}, \\
\E\!\left[k_\lambda(X_1,X_2)^2\ell_\mu(Y_1,Y_3)^2\right] &\lesssim \frac{1}{\LL}, \\
\E\!\left[k_\lambda(X_1,X_2)^2\ell_\mu(Y_3,Y_4)^2\right] &\lesssim \frac{1}{\LL},
\end{align*}
the constant in the notation `$\lesssim$' depends only on $d$ and $M$, where we recall that by assumption we have $\max\p{\|p_{xy}\|_\infty,\|p_{x}\|_\infty,\|p_{y}\|_\infty}\leq M$.
We deduce that
$$
\E\!\left[\Lambda^2_N\right] \lesssim \frac{1}{\LL}.
$$

As explained in \Cref{proof:fixed}, by relying on Dvoretzky–Kiefer–Wolfowitz inequality \citep{dvoretzky1956asymptotic,massart1990tight} as done by \citet[Proposition 4]{schrab2021mmd}, it is sufficient to prove upper bounds for the true permutation quantile $\widehat q_{1-\alpha}^{\,\lambda, \infty}$ without finite approximation.
From \Cref{eq:quantileperm}, we obtain that this quantile satisfies
$$
\widehat q_{1-\alpha}^{\,\lambda, \infty} \lesssim \max\p{
    \frac{\Lambda_N}{N}\ln\pp{\frac{1}{\alpha}},
    \frac{M_N}{N^{3/2}}\ln\pp{\frac{1}{\alpha}}^{3/2}
}.
$$
Using Markov's inequality and bounds obtained above, we get that
\begin{align}
    \widehat q_{1-\alpha}^{\,\lambda, \infty} &\lesssim
    \max\p{
        \frac{\sqrt{\mathbb{E}\left[\Lambda_N^2\right]}}{\sqrt{\delta}N}\ln\pp{\frac{1}{\alpha}},
        \frac{M_N}{N^{3/2}}\ln\pp{\frac{1}{\alpha}}^{3/2}
        }\nonumber\\
    \label{eq:quantileboundpermutation}
    \widehat q_{1-\alpha}^{\,\lambda, \infty} &\lesssim 
    \max\p{
        \frac{\ln\pp{{1}/{\alpha}}}{\sqrt{\delta}N\sqrt{\LL}},
        \frac{\ln\pp{{1}/{\alpha}}^{3/2}}{N^{3/2}\LL}
        }
\end{align}
holds with probability at least $1-\delta$ where $\delta\in(0,1)$.

Now, recall that by assumption ${4s>d}$ and 
$
\lambda_i^* = N^{-2/(4s+d)}
$
for $i=1,\dots,d$,
so that
$$
\frac{1}{\LLs}
= N^{{2 d}/{(4s+d)}}
< N
\qquad \Longleftrightarrow \qquad
N^{-1/2} < \sqrt{\LLs}
$$
which gives
\begin{align*}
    \widehat q_{1-\alpha}^{\,\lambda^*, \infty} &\lesssim
    \max\p{
        \frac{\ln\pp{{1}/{\alpha}}}{\sqrt{\delta}N\sqrt{\LLs}},
        \frac{\ln\pp{{1}/{\alpha}}^{3/2}}{N\sqrt{\LLs}}
        }\\
    &\lesssim \frac{\ln\pp{{1}/{\alpha}}^{3/2}}{\sqrt{\delta}N\sqrt{\LLs}}
\end{align*}
holding with probability at least $1-\delta$, since $\alpha\in\p{0,e^{-1}}$.
By combining this result with the reasoning of \citet{albert2019adaptive} as explained in \Cref{proof:fixed}, we obtain that the probability of type II error of the test is controlled by $\beta\in(0,1)$ when
$$ 
  \|{p-q}\|^2_2 ~\geq~ \|{(p-q)-T_{\lambda^*}(p-q)}\|^2_2 + C \frac{1}{N}\frac{\ln\!\p{{1}/{\alpha}}^{3/2}}{\beta \sqrt{\lambda_1^*\cdots\lambda_d^*}}.
$$
We have recovered the correct dependency with respect to $N$ and $\lambda$ with an improved $\alpha$-dependency of $\ln\pp{{1}/{\alpha}}^{3/2}$ compared to the $\alpha^{-1/2}$ dependency obtained by \citet[Proposition 8.7]{kim2020minimax}.
The proof of minimax optimality of the quadratic-time test with fixed bandwidth $\lambda^*$ does not depend on the $\alpha$-dependency and can be derived in both our setting and the one of \citet{kim2020minimax} using quantiles obtained from permutations by following the reasoning of \citet[Corollary 2]{albert2019adaptive}.
We obtain that the uniform separation rate over the Sobolev ball $\mathcal{S}_d^s(R)$ is, up to a constant,
$
    N^{-2s/(4s+d)}
$.
The improved $\alpha$-dependency is crucial for deriving the rate of the aggregated quadratic-time test over Sobolev balls because the weights appear in the $\alpha$-term (\emph{i.e.} $\alpha$ is replaced by $\alpha w_\lambda$ which depends on the sample size $N$).

\subsubsection{Proof of \Cref{theo:indminimax} (ii)}

Similarly to the proofs of \citet[Corollary 3]{albert2019adaptive} and \citet[Corollary 10]{schrab2021mmd}, consider
$$
\ell^* 
\coloneqq 
\Big\lceil\frac{2}{4s+d}\log_2\!\Big(\frac{N}{\ln(\ln(N))}\Big)\Big\rceil
\leq 
\Big\lceil\frac{2}{d}\log_2\!\Big(\frac{N}{\ln(\ln(N))}\Big)\Big\rceil
$$
and the bandwidth $\lambda^* \coloneqq \p{2^{-\ell^*},\dots,2^{-\ell^*}} \in \Lambda$ which satisfies
$$
\ln\p{\frac{1}{w_{\lambda^*}}}
\lesssim
\ln\p{\ell^*}
\lesssim
\ln(\ln(N))
$$
as $w_{\lambda^*} \coloneqq 6\pi^{-2}\p{\ell^*}^{-2}$, and
$$
\frac{1}{2}\p{\frac{N}{\ln\pp{\ln\pp{N}}}}^{-2/(4s+d)}
\leq
\lambda_i^*
\leq
\p{\frac{N}{\ln\pp{\ln\pp{N}}}}^{-2/(4s+d)}
$$
for $i=1,\dots,d$.
Since ${4s>d}$, we have
$$
\frac{1}{\LLs}
\leq 2^d\p{\frac{N}{\ln(\ln(N))}}^{{2 d}/{4s+d}}
\lesssim \frac{N}{\ln(\ln(N))}
\quad \Longleftrightarrow \quad
N^{-1/2} \lesssim \frac{\sqrt{\LLs}}{\sqrt{\ln(\ln(N))}}.
$$
By \Cref{eq:quantileboundpermutation}, we get that
\begin{equation}
    \label{eq:quantilemax}
    \widehat q_{1-\alpha w_{\lambda^*}}^{\,\lambda^*, \infty} 
    \lesssim
    \max\p{
        \frac{\ln\!\pbig{{1}/{(\alpha w_{\lambda^*})}}}{\sqrt{\delta}N\sqrt{\LLs}},
        \frac{\ln\!\pbig{{1}/{(\alpha w_{\lambda^*})}}^{3/2}}{N^{3/2}\LLs}
        }
\end{equation}
holds with probability at least $1-\delta$ for $\delta\in(0,1)$.
If the largest term in \Cref{eq:quantilemax} is the first one, then we get
\begin{align}
    \widehat q_{1-\alpha w_{\lambda^*}}^{\,\lambda^*, \infty} 
    &\lesssim 
    \frac{\ln\!\p{1/\alpha} + \ln\!\p{1/w_{\lambda^*})}}{\sqrt{\delta}N\sqrt{\LLs}}, \nonumber\\
    \label{eq:firstterm}
    \widehat q_{1-\alpha w_{\lambda^*}}^{\,\lambda^*, \infty} 
    &\lesssim
    \frac{\ln(\ln(N))}{\sqrt{\delta}N\sqrt{\LLs}},
\end{align}
where the constant in `$\lesssim$' also depends on $\alpha$.
The result then follows exactly as in the proofs of \citet[Corollary 3]{albert2019adaptive} and \citet[Corollary 10]{schrab2021mmd}.
So, we consider the case where the second term in \Cref{eq:quantilemax} is the largest one, so that
\begin{align*}
    \widehat q_{1-\alpha w_{\lambda^*}}^{\,\lambda^*, \infty} 
    &\lesssim 
    \frac{\ln\!\pbig{{1}/{(\alpha w_{\lambda^*})}}^{3/2}}{N^{3/2}\LLs}\\
    &\lesssim
    \frac{\ln\!\pbig{{1}/{ w_{\lambda^*}}}^{3/2}}{N\LLs} N^{-1/2}\\
    &\lesssim
    \frac{\ln(\ln(N))^{3/2}}{N\LLs} \frac{\sqrt{\LLs}}{\sqrt{\ln(\ln(N))}}\\
    &=
    \frac{\ln(\ln(N))}{N\sqrt{\LLs}}.
\end{align*}
We have recovered the same dependency as in \Cref{eq:firstterm} when considering the first term as the largest one, and the proof then follows exactly the ones of \citet[Corollary 3]{albert2019adaptive} and \citet[Corollary 10]{schrab2021mmd}.
We have treated both cases in \Cref{eq:quantilemax}, we conclude that the uniform separation rate over the Sobolev balls $\big\{\mathcal{S}_d^s(R):R>0, s>d/4\big\}$ of the quadratic-time aggregated test using a quantile obtained with permutations is (up to a constant)
$$
    \p{\frac{N}{\ln\pp{\ln\pp{N}}}}^{-2s/(4s+d)}.
$$

\end{document}